\documentclass[acmtog,authorversion=true]{acmart}

\usepackage{booktabs}
\usepackage{graphicx}
\usepackage{booktabs}
\usepackage{multirow}
\usepackage{bigdelim}
\usepackage{tabu}

\newcommand{\SubdivNet}{{SubdivNet}}
\definecolor{lightblue}{rgb}{0, 0, 0}

\def\zapcolorreset{\let\reset@color\relax\ignorespaces}
\def\colorrows#1{\noalign{\aftergroup\zapcolorreset#1}\ignorespaces}

\acmJournal{TOG}

\begin{document}
\title{Subdivision-Based Mesh Convolution Networks}

\author{Shi-Min Hu}
\orcid{0000-0001-7507-6542}
\email{shimin@tsinghua.edu.cn}
\author{Zheng-Ning Liu}
\orcid{0000-0001-6643-6016}
\email{lzhengning@gmail.com}
\author{Meng-Hao Guo}
\orcid{0000-0002-4128-4594}
\email{gmh20@mails.tsinghua.edu.cn}
\author{Jun-Xiong Cai}
\email{caijunxiong000@163.com}
\author{Jiahui Huang}
\email{huang-jh18@mails.tsinghua.edu.cn}
\author{Tai-Jiang Mu}
\email{taijiang@tsinghua.edu.cn}

\affiliation{%
 \institution{BNRist, Department of Computer Science and Technology, Tsinghua University}
 \city{Beijing}
 \country{China}}
\author{Ralph R. Martin}
\affiliation{%
 \institution{Cardiff University}
 \city{Cardiff}
 \country{UK}
}
\affiliation{%
 \institution{Tsinghua University}
 \state{Beijing}
 \country{China}
}
\email{MartinRR@cs.cf.ac.uk}

\begin{abstract}
Convolutional neural networks (CNNs) have made great breakthroughs in 2D computer vision. However, their irregular structure  makes it hard to harness the potential of CNNs directly on meshes.
A subdivision surface  provides a hierarchical multi-resolution structure, in which each face in a closed 2-manifold triangle mesh is exactly adjacent to three faces. Motivated by these two observations, this paper presents \emph{SubdivNet}, an innovative and versatile CNN framework for 3D triangle meshes with Loop subdivision sequence connectivity.
Making an analogy between mesh faces and pixels in a 2D image allows us to present a mesh convolution operator to aggregate local features from nearby faces. By exploiting face neighborhoods, this convolution can support standard 2D convolutional network concepts, e.g.\ variable kernel size, stride, and dilation. Based on the multi-resolution hierarchy, we make use of  pooling layers which uniformly merge four faces into one and an upsampling method which splits one face into four.
Thereby, many popular 2D CNN architectures can be easily adapted to process 3D meshes.
Meshes with arbitrary connectivity can be remeshed to have Loop subdivision sequence connectivity via self-parameterization, making SubdivNet a general approach.
Extensive evaluation and various applications demonstrate {\SubdivNet}'s effectiveness and efficiency. \end{abstract}

%
%


%
%


\begin{teaserfigure}
  \includegraphics[width=\textwidth]{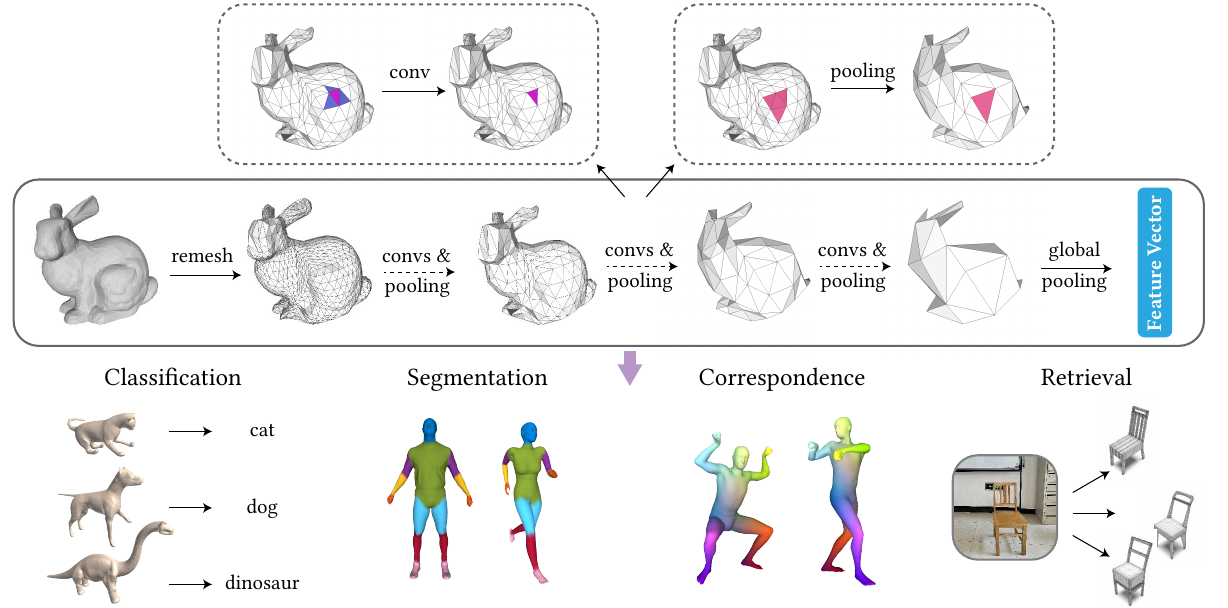}
  \caption{SubdivNet, a subdivision-based mesh convolution network for deep geometric learning. Given a mesh as input, we construct a hierarchical subdivision structure with a pyramid of regular connectivities, analogous to a 2D image pyramid. This structure permits natural notions of convolution, pooling, and upsampling operation on  3D meshes, which together provide the building blocks of our mesh-based deep neural network. Our network is effective and efficient for mesh-based representation learning in a variety of applications.
  }
  \label{fig:teaser}
\end{teaserfigure}

\maketitle

\section{Introduction}
Deep convolutional networks (CNNs) have achieved great success in 2D computer vision, leading to their generalization to a variety of disciplines, including 3D geometry processing. PointNet~\cite{DBLP:conf/cvpr/QiSMG17} is a pioneering approach for learning a feature representation of a point cloud; it has led to the development of more powerful networks~\cite{DBLP:conf/nips/QiYSG17,DBLP:conf/nips/LiBSWDC18}. 3D geometric learning has also been extended to other forms of 3D data, such as voxels~\cite{DBLP:conf/iccv/KlokovL17,DBLP:journals/tog/WangLGST17} and meshes~\cite{DBLP:journals/tog/HanockaHFGFC19,DBLP:journals/tog/LahavT20}.

In this paper, we consider 3D geometric learning using a mesh representation. Polygonal meshes are one of the most common 3D data representations, with applications in modeling, rendering, animation, 3D printing, etc. Unlike point clouds, meshes contain topological information. Polygon meshes can represent geometric context more effectively than voxels, since they only depict the boundaries of objects, omitting superfluous elements for the interior. 

Underpinning the success of 2D CNNs in the image domain is the inherently regular grid structure allowing hierarchical image pyramids, which enable CNNs to explore features of varying sizes by downsampling and upsampling.
However, an arbitrary mesh is irregular and lacks the gridded arrangement of pixels in images, making it difficult to define a standard convolution with variable kernel size, stride, and dilation. This prevents 3D geometric learning methods from taking advantage of mature network architectures used in the image domain. Furthermore, the unstructured connectivity between vertices and faces precludes finding a simple fine-to-coarse structure for meshes. One may consider levels of detail (LOD) via mesh simplification, but the mapping of geometry primitives between each level is not well-defined, and one cannot derive an intuitive pooling operator from a LOD hierarchy.

To apply the power of CNNs to meshes, efforts have been made to define special convolution operations on surfaces. Such methods typically attempt to encode or re-sample the neighborhood of a vertex into a regular local domain~\cite{DBLP:conf/iccvw/MasciBBV15,DBLP:conf/nips/BoscainiMRB16,DBLP:conf/cvpr/MontiBMRSB17,DBLP:journals/tog/MaronGATDYKL17,DBLP:conf/cvpr/TatarchenkoPKZ18,DBLP:journals/tog/PoulenardO18,DBLP:conf/cvpr/HuangZYFNG19}, where the convolutional operation can be derived.
Recently, mesh downsampling schemes have been proposed to dynamically merge regions using edge collapse~\cite{DBLP:journals/tog/HanockaHFGFC19,DBLP:conf/nips/MilanoLR0C20}, but they do not guarantee a uniformly enlarged receptive field everywhere as in 2D pooling.
Although the attention mechanism~\cite{DBLP:conf/iclr/VelickovicCCRLB18} can be applied to capture global context by treating a mesh as a graph, doing so comes at a heavy computational cost.

Instead, motivated by image pyramids in 2D CNNs that allow local features to be aggregated into larger-scale features at different levels, we note that subdivision surfaces also construct well-defined hierarchical mesh pyramids. A subdivision surface is a smooth surface produced by refining a coarse mesh.
In particular, in Loop subdivision~\cite{loop1987smooth}, each triangle mesh face is split into 4 triangles and then vertex positions are updated to smooth the new mesh (see Fig.~\ref{fig:4to1_mapping}(a)). As a result, the Loop subdivision scheme gives a 1-to-4 face mapping from the coarse mesh to the finer one. Correspondingly, if a mesh has the same connectivity as a Loop subdivision surface, it has a natural correspondence to a one-level-coarser mesh, and indeed, a carefully constructed Loop subdivision surface may preserve the Loop property over several levels leading to a fine to coarse hierarchy.
Furthermore, each face in any closed 2-manifold mesh is exactly surrounded by 3 other faces. The fixed number of face neighbors suggests a regular structure analogous to that of pixels in images,
 making it suitable for deriving a standard convolution operation on a mesh.

In this paper, we propose \emph{\SubdivNet}, which can learn feature representations for meshes with Loop subdivision sequence connectivity. Using the neighbors of faces, we define a novel convolution on mesh faces that supports variable kernel size, stride, and dilation. Thus, ours can operate on a large receptive field. Because of the flexibility of our new convolution on triangle meshes, successful neural networks
in the image domain, such as VGG~\cite{DBLP:journals/corr/SimonyanZ14a}, ResNet~\cite{ResNet16}, and DeepLabv3+~\cite{DBLP:conf/eccv/ChenZPSA18}, can be naturally adapted to meshes. 

{\SubdivNet} requires a mesh with subdivision sequence connectivity as input, which may appear excessively limiting. However, any triangle mesh representing a closed 2-manifold with arbitrary genus can be remeshed to have this property via self-parameterization~\cite{DBLP:conf/siggraph/LeeSSCD98,DBLP:journals/tog/LiuKCAJ20}. Thereby, {\SubdivNet} can be used as a general feature extractor for any closed 2-manifold triangle mesh.

{\SubdivNet} achieves state-of-the-art performance on 3D shape analysis, e.g.\ mesh classification, segmentation, and shape correspondence. Ablation studies verify the effectiveness of the proposed convolution, subdivision-based pooling, and advanced network architectures.

In summary, our work makes the following contributions:
\begin{itemize}
	\item a general mesh convolutional operation that permits variable kernel size, stride, and dilation analogous to standard 2D convolutions, making it possible to adapt well-known 2D CNNs to mesh tasks,
	\item {\SubdivNet}, a general mesh neural network architecture based on mesh convolution and subdivision sequence connectivity, with uniform pooling and upsampling, for geometric deep learning, supporting dense prediction tasks,
	\item demonstrations that {\SubdivNet} provides excellent results for various applications, such as shape correspondence and shape retrieval.
\end{itemize}

\section{Related Work}

\subsection{3D Geometric Learning}

One way of applying deep learning to geometric data is to transform 3D shapes into images, e.g.\ , an unordered set of projections~\cite{DBLP:conf/iccv/SuMKL15}, panoramas~\cite{DBLP:journals/spl/ShiBZB15}, or geometry images~\cite{DBLP:conf/eccv/SinhaBR16}, and then run 2D CNNs on them.
This family of \emph{indirect} methods is pose-sensitive because an additional view-dependent projection step is involved.
Another line of \emph{direct} solutions is to represent the shapes in their intrinsic 3D space, such as volumetric data, whereupon 3D CNNs can be applied~\cite{DBLP:conf/cvpr/WuSKYZTX15,DBLP:conf/iros/MaturanaS15} or adapted for higher resolution~\cite{DBLP:conf/iccv/KlokovL17,DBLP:journals/tog/WangLGST17,DBLP:journals/tvcg/LiuCKKH21}.
Recently, point-based learning techniques have emerged~\cite{DBLP:conf/cvpr/QiSMG17,DBLP:conf/nips/QiYSG17,DBLP:conf/nips/LiBSWDC18,DBLP:journals/tog/WangSLSBS19} due to the ease of acquisition of point cloud data by 3D sensors.
Nevertheless, the high computational demand for volumetric data and the absence of topological information for point clouds make current pipelines inefficient. However, methods that learn on \emph{mesh surfaces} overcome the above problems and have shown promise. Readers are referred to recent surveys~\cite{DBLP:journals/spm/BronsteinBLSV17,DBLP:journals/cvm/XiaoLZLG20} for a comprehensive review on 3D geometric learning.

\subsection{Deep Learning on Meshes}
Meshes are composed of three distinct types of geometric primitives: vertices, edges, and faces. We categorize mesh deep learning methods based on which of these are treated as the primary data.

\subsubsection{Vertex-based}
Certain works perform deep learning on 3D shapes by locally encoding  the neighborhood of sampled points into a regular domain, whereupon convolution operations (or kernel functions) can mimic those for images. Masci et al.~\shortcite{DBLP:conf/iccvw/MasciBBV15}, Boscaini et al.~\shortcite{DBLP:conf/nips/BoscainiMRB16}, Tatarchenko et al.~\shortcite{DBLP:conf/cvpr/TatarchenkoPKZ18}, and Poulenard et al.~\shortcite{DBLP:journals/tog/PoulenardO18} parameterize geodesic patches into 2D domains, e.g.\ the tangent plane, for use with 2D CNNs or point networks. TextureNet~\cite{DBLP:conf/cvpr/HuangZYFNG19} and PFCNN~\cite{DBLP:conf/cvpr/YangLPLT20} extend geodesic convolution by better handling inconsistent orientations of tangent spaces. Global parameterization is employed by Maron et al.~\shortcite{DBLP:journals/tog/MaronGATDYKL17} and Haim et al.~\shortcite{DBLP:conf/iccv/HaimSBML19} to perform surface convolution. Such methods are usually insensitive to the meshing of shapes due to parameterization. Another series of works applies convolution directly on the mesh structure. Some approaches ~\cite{DBLP:conf/cvpr/MontiBMRSB17,9184119,DBLP:conf/cvpr/DaiN19,DBLP:conf/cvpr/KostrikovJPZB18} employ graph neural networks (GNN) to use vertex connectivity. Pixel2Mesh~\cite{DBLP:conf/eccv/WangZLFLJ18} generates meshes from coarse to fine via subdivision, and updates geometry by GNN. Lim et al.~\shortcite{DBLP:conf/eccv/LimDCK18} and Gong et al.~\shortcite{,gong2019spiralnet++} propose a spiral convolution pattern within the $k$-ring neighborhood of a vertex. DiffusionNet~\cite{sharp2021diffusionnet} and HodgeNet~\cite{DBLP:journals/tog/0001021} extend the Laplacian operator to learn the surface representation. While these methods can learn the local representation, they are usually less capable of learning multi-scale and contextual information in a mesh.

Closer to our approach are methods with hierarchical design. Dilated kernel parametrization~\cite{DBLP:conf/cvpr/YiSGG17}, and mesh downsampling and upsampling~\cite{DBLP:conf/eccv/RanjanBSB18} can be adopted in the spectral domain to define mesh convolution to aggregate multi-scale information.
Schult et al.~\shortcite{DBLP:conf/cvpr/SchultEKL20} combine two kinds of convolutions separately defined on neighbors according to geodesic and Euclidean distance, also exploiting mesh simplification to provide a multi-resolution architecture.
Based on subdivision sequence connectivity, our approach offers a more general and standard convolution directly defined on the mesh; as such, it supports variable kernel size, stride, and dilation.

\subsubsection{Edge-based}
Each edge in a 2-manifold triangle mesh is adjacent to two faces and  four `next' edges.
This property is exploited by MeshCNN~\cite{DBLP:journals/tog/HanockaHFGFC19} to define an ordering invariant convolution. 
PD-MeshNet~\cite{DBLP:conf/nips/MilanoLR0C20} constructs a primal graph and a dual graph from the input mesh, then performs convolutions on these graphs using a graph attention network~\cite{DBLP:journals/corr/abs-1806-00770,DBLP:conf/iclr/VelickovicCCRLB18}. MeshWalker~\cite{DBLP:journals/tog/LahavT20} extracts shape features by  walking along edges rather than exploiting neighborhood structures.

MeshCNN and PD-MeshNet dynamically contract edges to simplify meshes within the network, whereas MeshWalker build the hierarchy using variable-step walks. Unlike our approach, they do not provide a downsampling scheme to uniformly expand the receptive field, a successful strategy in 2D CNNs.

\subsubsection{Face-based} Face based methods focus on how to efficiently and effectively gather information from neighboring faces.
Xu et al.~\shortcite{DBLP:conf/iccv/XuDZ17} propose a rotationally-invariant face based method considering $k$-ring neighbors for defining convolution on meshes; it is guided by face curvature. MeshSNet, proposed by Lian et al.~\shortcite{DBLP:conf/miccai/LianWWLDKS19}, adopts graph-constrained mesh-cell nodes to integrate local-to-global geometric features. MeshNet~\cite{DBLP:conf/aaai/FengFYZG19} learns the spatial and structural features of a face by aggregating its 1-ring neighbors with the help of two mesh convolutional layers. DNF-Net~\cite{DBLP:journals/corr/abs-2006-15510} uses multi-scale embedding and a residual learning strategy to denoise mesh normals on cropped local patches.  
Hertz et al.~\shortcite{DBLP:journals/tog/HertzHGC20} generate geometric textures, using a 3-face convolution and a subdivision-based upsampling similar to Pixel2Mesh~\cite{DBLP:conf/eccv/WangZLFLJ18}.

{\SubdivNet} uses a regular and uniform downsampling scheme to establish a fine-to-coarse mesh hierarchy. Our convolution also exploits distant neighbors of faces and thus can have a larger receptive field. Compared to Xu et al.~\shortcite{DBLP:conf/iccv/XuDZ17}, our convolutions efficiently support stride and large dilation, allowing us to better capture long-range features.

\subsection{Subdivision Surfaces and Multiresolution Modeling}
A subdivision surface is a smooth surface produced by refining a coarse mesh. The best-known mesh subdivision algorithms are  Catmull-Clark subdivision~\cite{CatmullClark78} for quad meshes and Loop subdivision~\cite{loop1987smooth} for triangle meshes. They insert new vertices and edges, split faces, and linearly update vertex positions. Other subdivision schemes, e.g.\ \cite{doo1978behaviour,DBLP:journals/tog/DynLG90,DBLP:conf/siggraph/Kobbelt00} and non-linear approaches, e.g.\ \cite{DBLP:journals/tog/LiuPWYW06,DBLP:journals/cagd/SchaeferVG08} have also been proposed.

Multi-resolution modeling, also known as level of detail, aims to construct a sequence of meshes from fine to coarse, and is widely applied in mesh compression, editing, and fast rendering. 
There have been many works on this topic, and we will only consider methods that maintain subdivision sequence connectivity from fine to coarse: we need a whole hierarchy of meshes \emph{all} having Loop subdivision connectivity.
MAPS~\cite{DBLP:conf/siggraph/LeeSSCD98} is a pioneering work that computes a parameterization of a mesh over a simplified version of the mesh. Then, a new mesh with subdivision sequence connectivity is constructed on the surface of the simplified mesh. Finally, the vertices of the new mesh are projected back to  the input faces via the parameterization.
 This idea is further improved in terms of distortion and smoothness by~\cite{DBLP:journals/cgf/KobbeltVLS99,DBLP:conf/siggraph/GuskovVSS00,DBLP:conf/compgeom/GuskovKSS02,DBLP:journals/tog/KhodakovskyLS03}. 
Liu et al~\shortcite{DBLP:journals/tog/LiuKCAJ20} have also extended the MAPS algorithm to generate multi-resolution meshes for network training.

By use of such a multi-resolution method, any mesh can be remeshed to have subdivision sequence connectivity, making SubdivNet a general method for 3D mesh analysis.

\section{{\SubdivNet}}
\subsection{Notation}

\begin{figure}[t!]
    \centering
    \includegraphics[width=\linewidth]{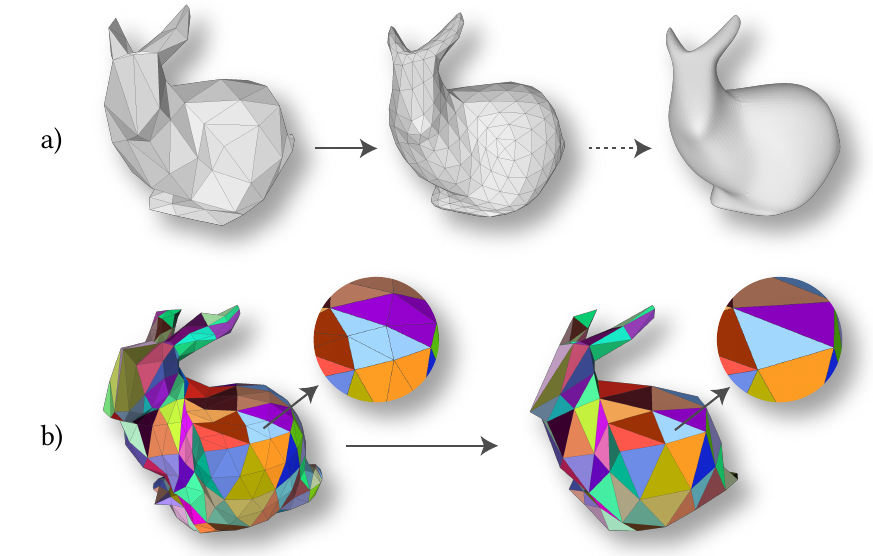}
    \caption{(a) Loop subdivision. A coarse mesh is iteratively refined by splitting each face into four and smoothing. (b) The 4-to-1 face mapping from a fine mesh to the next coarser level using Loop subdivision sequence connectivity.}
    \label{fig:4to1_mapping}
\end{figure}

Before giving details of {\SubdivNet}, it is necessary to define the mathematical notation used throughout this paper.

A triangle mesh $\mathcal{M} = (\mathbf{V}, \mathbf{F})$ is defined by a set of vertices $\mathbf{V} = \{ v_i | v_i \in \mathcal{R}^3\}$ and a set of triangular faces $\mathbf{F} = \{f_i | f_i \in \{1, \cdots, |\mathbf{V}| \}^3\}$, indicating the triangle's vertices, and hence implicitly, the connectivity. Each face $f_i$ holds an input feature vector $e_i$, which is to be processed by {\SubdivNet}. 

Two faces $f_i$ and $f_j$ are said to be adjacent if they share an edge. The distance $D(f_i, f_j)$ between $f_i$ and $f_j$ is defined as the minimum number of faces traversed by any path from one to the other across edges. The $k$-ring neighborhood of $f_i$ is then:
\[
    \mathcal{N}_k(f_i) = \{ f_j | D(f_i, f_j) = k \}.
\]

We say that a triangle mesh $\mathcal{M}$ has \emph{Loop subdivision connectivity} if it has the same connectivity as a mesh formed by one round of Loop subdivision acting on a coarser mesh.

We say that a triangle mesh $\mathcal{M}$ has \emph{Loop subdivision sequence connectivity} if there exists a sequence of meshes $(\mathcal{M}_0, \cdots, \mathcal{M}_L)$, $L \geq 1$, 
where $\mathcal{M}_L=\mathcal{M}$,
satisfying two requirements: (i) all except possibly $\mathcal{M}_0$ have Loop subdivision connectivity; (ii) all vertices in $\mathcal{V}_{i}$ are also present in $\mathcal{V}_{i+1}$, $0<i<L$. The bunny in Fig.~\ref{fig:teaser} illustrates such a sequence.

We refer to $L$ as the \emph{subdivision depth} of $\mathcal{M}$, 
$\mathcal{M}_0$ as the \emph{base mesh} of $\mathcal{M}$, and the number of faces of $\mathcal{M}_0$ as the \emph{base size}. Clearly, the number of faces of $\mathcal{M}$ is $|\mathbf{F}| = 4^L |\mathbf{F}_0|$.

If $\mathcal{M}$ has Loop subdivision sequence connectivity, we can establish a 4-to-1 face mapping from each mesh $\mathcal{M}_i$, $i>0$,  to  mesh $\mathcal{M}_{i-1}$, which can be regarded as the topological inverse of Loop subdivision (i.e.\ ignoring vertex geometry updates) (see Fig.~\ref{fig:4to1_mapping}(b)).

Most common meshes, whether designed by artists or scanned by sensors, lack Loop subdivision sequence connectivity. Hence, we first remesh the input mesh via a self-parameterization to confer this property on it, for a specified base size and subdivision depth, using methods explained in Sec.~\ref{sec:meshgen}.
The remainder of this section will assume that the input mesh has been appropriately remeshed. 

\subsection{Overview}

Given a watertight 2-manifold triangle mesh with Loop subdivision sequence connectivity, we aim to learn a global representation for the 3D shape, or feature vectors on each face for local geometry.

Like a 2D image pyramid, $(\mathcal{M}_L, \cdots, \mathcal{M}_0)$ provides a hierarchical structure, or \emph{mesh pyramid}. Through the 4-to-1 face mapping provided by subdivision sequence connectivity, we can also establish an injection of faces from $\mathcal{M}_L$ to $\mathcal{M}_0$ step by step, allowing feature aggregation from local to global. 

Based on this, {\SubdivNet} takes a mini-batch of closed 2-manifold triangle meshes with subdivision sequence connectivity as input.
It computes features with convolutions defined on triangle faces, and aggregates long-range feature descriptions of meshes by uniform face downsampling. Because of the regular number of face neighbours, the mesh convolution also supports variable kernel size, dilation, and stride, acting like a standard convolution in the image domain. To support dense prediction tasks, e.g.\ segmentation, an upsampling operation is provided as the inverse of pooling.

Due to the flexibility and generality of convolution, pooling and upsampling, we can directly adapt  well-known networks from the 2D image domain to mesh learning tasks.

\begin{figure*}[t!]
  \centering
  \includegraphics[width=\linewidth]{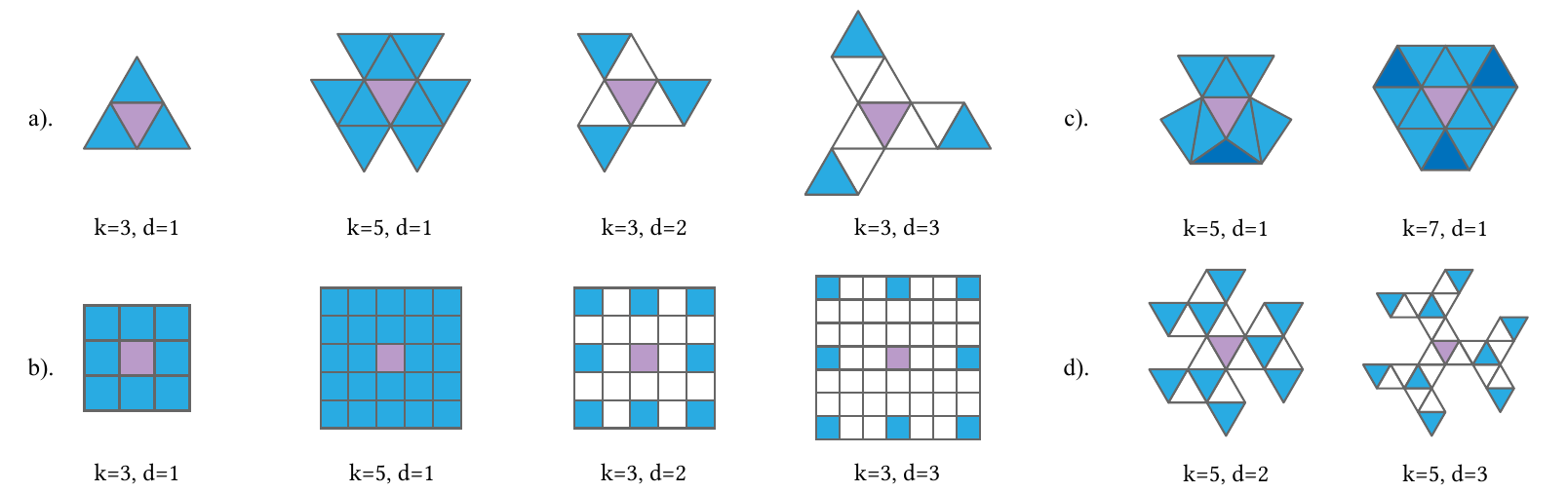}
  \caption{Mesh convolution kernel pattern. (a) Mesh convolution kernels with different kernel size $k$ and dilation $d$. (b) Corresponding 2D convolution kernels. (c) Duplication happens for kernel sizes larger than $3$: deep blue faces are accessed twice. (d) More complex convolution kernels with larger kernel size and dilation.}
  \label{fig:mesh_convs}
\end{figure*}

\subsection{Convolution}
This section discusses which faces should be considered in convolution, i.e.\ the convolution kernel pattern $\Omega$. Then we will discuss how to perform the convolution with such kernels patterns.

\subsubsection{Basic Convolution Pattern}
A key to defining convolution for a given signal is to specify its neighborhood, or the kernel pattern.
Since there are no boundary edges in a 2-manifold triangle mesh,
 each face on the mesh has exactly 3 adjacent faces. 
This 3-regular property is analogous to the lattice connectivity of pixels in 2D images, motivating us to define a basic convolution over faces. Formally, for each face $f_i$, the basic convolution kernel pattern is formed by its 1-ring neighbors $\Omega(f_i) = \mathcal{N}_1(f_i)$, illustrated in Fig~\ref{fig:mesh_convs}(a).

\subsubsection{Kernel Size}
To enable the convolution to have an larger receptive field, convolution in 2D images is designed to support a variable kernel size. This is also critical in shape analysis to learn more discriminative representations for each vertex and face, facilitating tasks such as shape correspondence and segmentation. We consider additional nearby faces to define the pattern of a convolution with variable kernel size $k$, so
\begin{equation}
\Omega(f_i, k) = {\bigcup\limits_{i=1}^{\hat{k}} \mathcal{N}_{\hat{k}}} (f_i), \quad \hat{k} = \frac{k-1}{2}, k=1,3,5,\dots
\end{equation}

In total, there are $3 \times (2^{\hat{k}}-1)$ faces in the kernel pattern. For instance, Fig.~\ref{fig:mesh_convs}(a) depicts a case with a kernel size of 5. 

However, when $k$ is greater than 3, adjacent faces may be counted more than once, resulting in  fewer  triangles in $\Omega$ than expected, as seen Fig.~\ref{fig:mesh_convs}(c).
Avoiding such issues would lead to a complex convolution design,
so we simply preserve all duplications of faces and keep $|\Omega|$ fixed. Aside from simplicity, another reason is that in modern networks, a larger 2D convolution kernel is usually substituted by a stack of small kernels. 
When $k = 3$, no duplication occurs. When $k = 5$, duplication can only exist around vertices whose degree is 4 or less (the degree of most vertices is 6 due to subdivision sequence connectivity). When the kernel size is larger than 7, faces may be accessed more than twice.

\subsubsection{Dilation}
Dilated convolution, also known as atrous convolution, is a widely used variation, where holes are inserted in the kernel pattern (see Fig~\ref{fig:mesh_convs}(b)). Dilation is an efficient strategy to expand the receptive field without consuming more computing or memory resources. To extend this concept to triangle meshes, we define that in a kernel pattern with dilation $d$, the distance between a face and its nearest face (including the center $f_i$) is $d$. In particular, the dilation of the basic convolution pattern is 1.
\begin{figure}
    \centering
    \includegraphics[width=\linewidth]{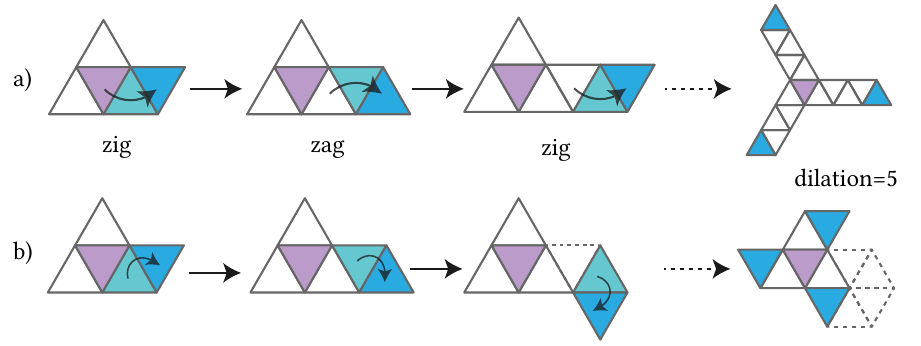}
    \caption{(a) Zig-zag strategy provides a uniformly dilated convolution from the basic convolution pattern.
    (b) Alternative scheme folds back, causing the dilation pattern to select triangles too close to the central triangle. }
    \label{fig:dilated}
\end{figure}
In a 2D image grid, the dilated kernel can be easily obtained by skipping rows and columns. However, such a strategy cannot be readily applied to triangle meshes (see Fig.~\ref{fig:dilated}(b)). 
Instead, we propose a \emph{zig-zag} strategy to define a kernel pattern with dilation $d$, shown in Fig~\ref{fig:dilated}(a). Taking a dilated convolution whose kernel size is 3 as an example, we move from faces in $\Omega(f_i, d)$ to  neighbors in turn $d$ times, alternately clockwise or counter-clockwise  with respect to the last position. Without loss of generality, if we assume that the input ordering of the three vertices in a face is counter-clockwise, then `zig' and `zag' are counter-clockwise and clockwise, respectively. Using the opposite definition is also feasible, which leads to a different, but symmetric pattern.

For kernel sizes $k$ greater than 3, theoretically, we may first find the $k$-ring neighbors and then perform dilation before the finding $(k+1)$-ring neighbors (see Fig.~\ref{fig:mesh_convs}(d)).

This is not the only way to define dilation, but this formulation is based on face distance, consistently with our approach to kernel size. Another motivation is to ensure that $|\Omega(f_i, d)| = |\Omega(f_i, d=1)|$ as required in 2D image grids. Furthermore, the zig-zag style results in a uniform spatial distribution of elements in $\Omega$, reducing the occurrence of duplicated faces. 

One may notice that the proposed dilation is asymmetric. As a result, only elements from three directions are considered when the kernel size is 3 while the information from the other three directions is lost, possibly leading to bias. However, with two or more dilated convolutions, features for all directions can be aggregated, avoiding potential bias.

\subsubsection{Stride}
In 2D CNNs, stride determines how densely the convolution is applied to the image. Because a 2D convolution with a stride greater than 1 reduces the resolution of the 2D feature map, it is frequently  used for downsampling, acting as a pooling layer with parameters. 

Thus, we also define a strided convolution based on the mesh pyramid. When the mesh convolution has a stride, it is only applied to the central face of a group of faces that merge into a single face in the coarser mesh. Since  1-to-4 Loop subdivision is used throughout {\SubdivNet}, the stride size is 2. To support an arbitrary stride size $s$, one can choose a 1-to-$s^2$ split subdivision scheme rather than the Loop subdivision scheme when remeshing the input. See Sec.~\ref{sec:meshgen} for further discussion of remeshing.

\subsubsection{Order-invariant Convolution Computation}
The three neighbors of a face are unordered, yet a robust convolution should be ordering invariant. While $\Omega(f_i)$ is an unordered set, we rearrange the set counter-clockwise around $f_i$, resulting in a sequence $\hat{\Omega}(f_i)$. Therefore, $\hat{\Omega}(f_i)$ is a closed ring (see Appendix~\ref{app:network} for details). Even so, where ring ordering starts is still ambiguous, but we can remove the ambiguity by computing order-invariant intermediate features. The convolution on a face $f_i$ is defined as, 
\begin{equation}
    \mathrm{Conv}(f_i) = w_0 e_i + w_1 \sum_{j=1}^{n} e_j + w_2 \sum_{j=1}^{n} |e_{j+1} - e_j| + w_3 \sum_{j=1}^{n} |e_i - e_j|,
\end{equation}
where $n = |\hat{\Omega}(f_i)|$. $e_j$ ($e_{n+1} = e_1$) is the feature vector on the $j$th face in $\hat{\Omega}(f_i)$ , and $(w_0,w_1,w_2,w_3)$ are learnable parameters. As summation is ordering invariant, the convolution is also insensitive to face ordering.

\subsection{Pooling}
With the pyramid of input meshes, pooling on triangle meshes is as simple as on a regular 2D image grid, as shown in Fig.~\ref{fig:pool_upsample}:  four subdivided faces in the finer mesh are pooled to the parent face in the coarser mesh. 

MeshCNN~\cite{DBLP:journals/tog/HanockaHFGFC19} defines pooling via dynamic edge collapse. To maintain a half-edge data structure, edge collapse is executed in sequence, while our pooling approach can be implemented in parallel. On the other hand, edge pooling cannot guarantee that all edges are downsampled once, but our uniform scheme ensures that all faces are involved in pooling. As a result, our approach produces a more spatially-uniform coarse mesh and achieves the same level of feature aggregation everywhere, though the face areas after pooling are not the same.

Like the convolution stride, the pooling operation supports a stride $s$ greater than 2 via a $1$-to-$s^2$ subdivision scheme when preprocessing the input meshes.

\subsection{Upsampling}
\begin{figure}[t!]
    \centering
    \includegraphics[width=\linewidth]{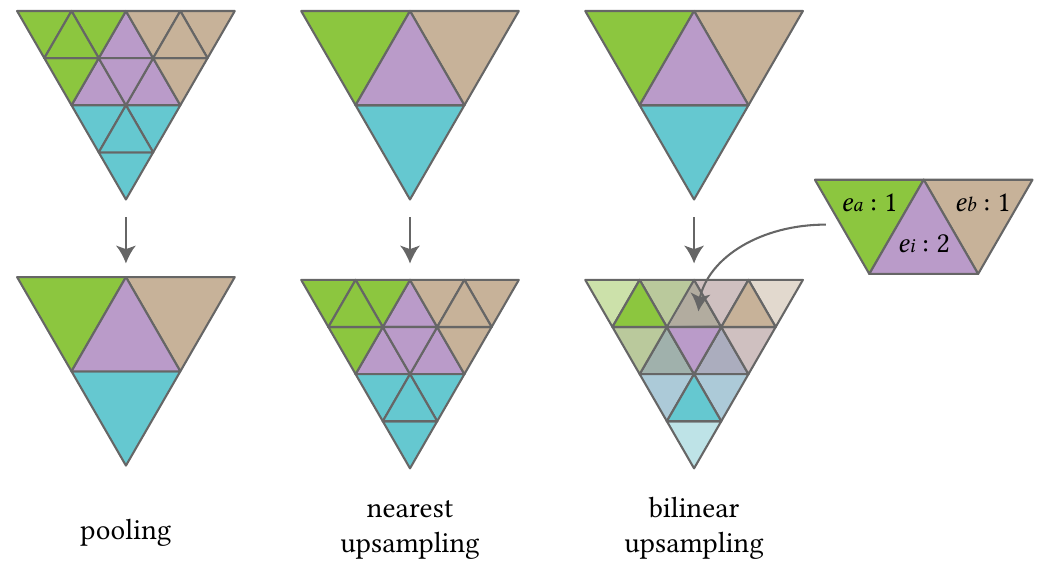}
    \caption{Pooling and upsampling}
    \label{fig:pool_upsample}
\end{figure}

As the reverse of pooling, upsampling is also defined with the help of the mesh pyramid. \emph{Nearest upsampling} simply splits a face into 4 faces; features on split faces are copied from the original face, as shown in Fig.~\ref{fig:pool_upsample}. 

In 2D dense prediction tasks, bilinear upsampling is also widely utilized. Therefore, we also define a \emph{bilinear upsampling} based on face distance, to provide smoother interpolation than nearest upsampling. See Fig.~\ref{fig:pool_upsample}. The feature of the central subdivided face $f_i$ is equal to the original face's feature, while the feature of the other faces $f_j$ is computed by, 
\begin{equation}
    e_j = \frac{1}{2}e_i + \frac{1}{4}e_{a} + \frac{1}{4}e_{b},
\end{equation}
where $f_a$ and $f_b$ are the other two faces adjacent to $f_j$.
\begin{figure*}[t!]
    \centering
    \includegraphics[width=\linewidth]{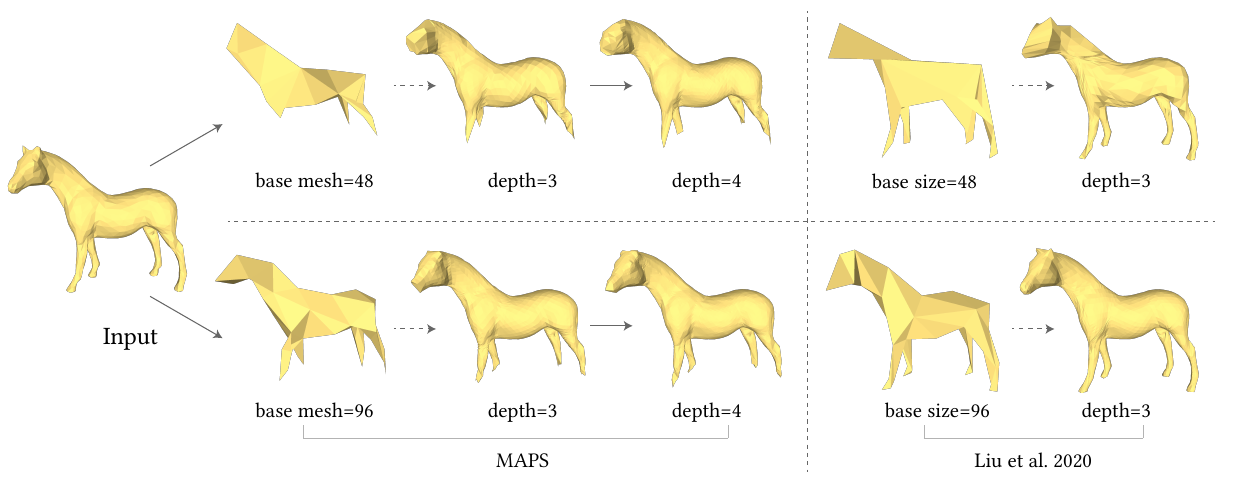}
    \caption{Remeshing results of MAPS and Liu et al.~\shortcite{DBLP:journals/tog/LiuKCAJ20}. MAPS first decimates the input mesh, and constructs a bijection between the original mesh and the base mesh. Then the base mesh is subdivided and new vertices are projected back onto the input. Limited sampling in the base mesh leads to obvious distortion, e.g.\ in the limbs. Liu et al.'s approach produces results of higher quality.}
    \label{fig:remeshing}
\end{figure*}

\subsection{Propagating Features to Raw Meshes}
Because the proposed convolution (without stride) is based on neighborhoods, it can also be applied on  raw input meshes  lacking subdivision sequence connectivity. We thus propose a \emph{feature propagation layer} to transfer information from the remeshed shape to the original mesh. This layer is similar to $k$-NN propagation in point cloud methods~\cite{DBLP:conf/nips/QiYSG17}. For each face $f^{R}$ in the raw mesh, we find the nearest face $f^{r}_{0}$ in the remeshed shape. Let $f^{r}_i$ ($i=1,2,3$) be the three adjacent faces of $f^{r}_0$. Then the feature vector on $f^R$ is obtained by interpolation, weighted by distance:
\begin{equation}
    e^{R} = {\sum_{i=0}^{3} \lambda_i e^r_i}/{\sum_{i=0}^{3} \lambda_i},\qquad \lambda_{i} = 1/{\mathbb{D}(f^R, f^r_i)^2},
\end{equation}
where $\mathbb{D}(\cdot)$ is the Euclidean distance between  face centers. 

With feature propagation and further convolution on the raw input, {\SubdivNet} can be end-to-end, making it convenient to incorporate other learning approaches on meshes. The extra layers can also aid in obtaining per-face predictions on raw meshes. However, it is simpler and more practical to project results from the remeshed shape onto the raw mesh by finding the nearest face. Therefore, feature propagation is not used in most of our experiments.

\subsection{Network Architecture}
Benefiting from the regularity that the proposed convolution offers, we can easily apply popular 2D convolutional networks to 3D meshes. In our experiments, we implement a VGG-style~\cite{DBLP:journals/corr/SimonyanZ14a} network for classification, and U-Net~\cite{DBLP:conf/miccai/RonnebergerFB15} and DeepLabv3+~\cite{DBLP:conf/eccv/ChenZPSA18} with a ResNet50~\cite{ResNet16} backbone for dense prediction. DeepLabv3+ provides state-of-the-art performance for 2D image segmentation; its key idea is to extend the receptive field by using dilated convolutions. This mechanism is also helpful for 3D meshes; we provide  network details in Appendix \ref{app:network}.

Previous work~\cite{DBLP:journals/tog/HanockaHFGFC19,DBLP:conf/nips/MilanoLR0C20} has adopted residual design and skip connections in networks, while not including all components of the original ResNet or DeepLabv3+, such as the strided convolution or the dilated convolution.

\subsection{Input features}
The input feature for each triangular face is a 13-dimensional vector, composed of a 7-dimensional shape descriptor and a 6-dimensional pose descriptor. The components of the shape descriptor are the face area, the three interior angles of the triangle, and the inner products of the face normal with the three vertex normals (characterizing curvature). The pose descriptor gives the position of the face center and the face normal, helping the network to identify faces with similar shapes through position and orientation. Further user-defined features such as color could also be added for specific learning tasks.

\section{Remeshing for Subdivision Connectivity}
\label{sec:meshgen}

{\SubdivNet} requires meshes with Loop subdivision sequence connectivity as input; however, most available meshes lack this property. We thus must remesh the input to have this property beforehand. 

\begin{figure}[t!]
    \centering
    \includegraphics[width=\linewidth]{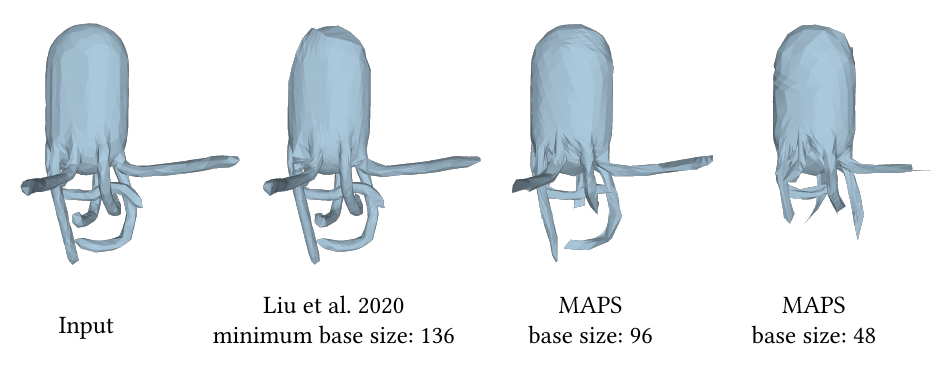}
    \caption{Octopus mesh from the SHREC11 dataset~\cite{DBLP:journals/tog/WangAK0CC12}. Liu et al.'s~\shortcite{DBLP:journals/tog/LiuKCAJ20} method cannot produce a lower base mesh size than 136 triangles. However, MAPS can reduce the base size while keeping important features. For example, although shortened, the number of tentacles is unchanged. Here, subdivision depth = 3.}
    \label{fig:failure_liu}
\end{figure}

One solution is to use a self-parameterization method, e.g.\ the MAPS algorithm~\cite{DBLP:conf/siggraph/LeeSSCD98} or an improved version~\cite{DBLP:journals/tog/LiuKCAJ20}. The key idea is to establish a mapping between the input mesh and the simplified mesh (the base mesh). Thus, by subdividing the base mesh and  back-projection onto the input mesh,  we can remesh the input with subdivision sequence connectivity. The output of MAPS occasionally exhibits visible distortion. Liu et al. improve the quality of output but cannot reach a low base size when the input mesh is complicated (see Fig.~\ref{fig:failure_liu}). See Appendix~\ref{app:remeshing} for further details.

In practice, we adopt a simple strategy to switch between remeshing approaches according to task. For tasks where the global shape is more important and local distortion can be tolerated, such as classification, we choose the MAPS algorithm to achieve a higher degree of global feature aggregation, whereas for tasks where local details are more crucial, e.g.\ fine-grained segmentation, we use Liu et al.'s approach and a larger base size. 

Both remeshing algorithms require the input to be manifold and closed, else local parameterizations may fail. More general meshes must be converted to watertight manifolds beforehand, via additional preprocessing (see Appendix~\ref{app:manifold40}).

\section{Experiments}

The generality and flexibility of our mesh convolution permits {\SubdivNet} to be applied to a wide range of 3D shape analysis tasks. 
We have quantitatively evaluated {\SubdivNet} for mesh classification, mesh segmentation, and shape correspondence, comparing it to state-of-the-art alternatives. 
We have conducted other qualitative experiments to demonstrate its applicability in other areas, such as real-world mesh retrieval. The key components of SubdivNet are also evaluated in an ablation study.

\subsection{Data preprocessing and augmentation}

As the meshes in the datasets do not have Loop subdivision sequence connectivity, we first remeshed all data in both the training and the test sets. As some datasets are small, we generated multiple remeshed meshes for each input by randomly permuting the order of vertex removal or edge collapse. Augmentation makes the network insensitive to remeshing. 

To reduce network sensitivity to size, we scaled each input to fit inside a unit cube and then applied random anisotropic scaling with a normal distribution $\mu = 1$ and $\sigma = 0.1$, following \cite{DBLP:journals/tog/HanockaHFGFC19}: for example, certain human body shapes are taller or thinner than others. 
We also find that some orientations of shapes in the test dataset do not appear in the training dataset, e.g.\ in the human body dataset~\cite{DBLP:journals/tog/MaronGATDYKL17}. 
Therefore, for such datasets, we also randomly changed the orientation of the input data by rotating it around the three axes with Euler angles of $0$, $\pi/2$, $\pi$, or $3\pi/2$.

\subsection{Classification}

We first demonstrate the capabilities of {\SubdivNet} for mesh classification  using three datasets. As in the data augmentation adopted during training, we additionally generated 10 to 20 remeshed shapes for each mesh in the test, and a majority voting strategy was applied to reduce the variance introduced by remeshing.

\subsubsection{SHREC11}
The SHREC11 dataset~\cite{DBLP:conf/3dor/LianGBDHKKLNOOORSSSTV11} contains 30 classes with 20 samples per class. Following the setting in~\cite{DBLP:journals/tog/HanockaHFGFC19}, {\SubdivNet} is evaluated using two protocols with 16 or 10 training examples in each class. We report the average accuracy on 3 random splits into training and test sets in Table~\ref{tab:shrec}. {\SubdivNet} correctly classifies all test meshes. 
Even without  majority voting, {\SubdivNet} is still comparable to or outperforms other state-of-the-art methods. With voting, accuracy already reaches 100\% when accuracy without voting is around 95\% in training. This suggests that the proposed method is sufficiently accurate  for the SHREC11 mesh classification task.

\begin{table}[t!]
    \centering
    \begin{tabu}{lrr}
    \toprule
        Method                                          & Split 16  & Split 10  \\ 
    \midrule
        GWCNN~\cite{DBLP:journals/cgf/EzuzSKB17}        & 96.6\%    & 90.3\%    \\
        MeshCNN~\cite{DBLP:journals/tog/HanockaHFGFC19} & 98.6\%    & 91.0\%    \\
        PD-MeshNet~\cite{DBLP:conf/nips/MilanoLR0C20}   & 99.7\%    & 99.1\%    \\
        MeshWalker~\cite{DBLP:journals/tog/LahavT20}    & 98.6\%    & 97.1\%    \\
        HodgeNet~\cite{DBLP:journals/tog/0001021}       & 99.2\%    & 94.7\%    \\
        DiffusionNet~\cite{sharp2021diffusionnet}       & -         & 99.7\%    \\
        SubdivNet (w/o majority voting)                 & \textbf{99.9}\%    & \textbf{99.5}\% \\
        SubdivNet                                       & \textbf{100}\% & \textbf{100}\% \\ 
    \bottomrule
    \end{tabu}
	\vspace{0.5em}
    \caption{Classification accuracy on the SHREC11 dataset~\cite{DBLP:conf/3dor/LianGBDHKKLNOOORSSSTV11}. We believe that SubdivNet is the first method to achieve  perfect performance.}
    \label{tab:shrec}
\end{table}

\subsubsection{Cube Engraving}

The Cube Engraving dataset~\cite{DBLP:journals/tog/HanockaHFGFC19} was synthesized by engraving 2D shapes on one random face of a cube. There are 22 categories and 4,381 shapes in the released dataset. {\SubdivNet} is the first method to make no mistakes, as shown in Table~\ref{tab:cube}.

\begin{table}[t!]
    \centering
    \begin{tabular}{lr}
    \toprule
        Method                                              & Accuracy \\
    \midrule
        PointNet++~\cite{DBLP:conf/nips/QiYSG17}            & 64.3\%\\
        MeshCNN~\cite{DBLP:journals/tog/HanockaHFGFC19}     & 92.2\% \\
        PD-MeshNet~\cite{DBLP:conf/nips/MilanoLR0C20}       & 94.4\% \\
        MeshWalker~\cite{DBLP:journals/tog/LahavT20}        & 98.6\% \\
        SubdivNet (w/o majority voting)                     & \textbf{98.9}\% \\
        SubdivNet                                           & \textbf{100.0}\% \\
    \bottomrule
    \end{tabular}
	\vspace{0.5em}
    \caption{Classification accuracy on the Cube Engraving dataset~\cite{DBLP:journals/tog/HanockaHFGFC19}. SubdivNet is the first method to correctly classify all test meshes. }
    \label{tab:cube}
\end{table}

\subsubsection{Manifold40}
ModelNet40~\cite{DBLP:conf/cvpr/WuSKYZTX15}, containing 12,311 shapes in 40 categories, is a widely used benchmark for 3D geometric learning. However, most  3D shapes in ModelNet40 are not watertight or 2-manifold, leading to remeshing failures. Therefore, we reconstructed the shapes in ModelNet40 and built a corresponding \emph{Manifold40} dataset, in which all shapes are closed manifolds. See Appendix~\ref{app:manifold40} for details of Manifold40 and the specific experimental settings.

We trained and evaluated point cloud methods, MeshNet~\cite{DBLP:conf/aaai/FengFYZG19}, MeshWalker~\cite{DBLP:journals/tog/LahavT20}, and {\SubdivNet} on Manifold40; the results are shown in Table~\ref{tab:modelnet40}. Because of the reconstruction error and simplification distortion, Manifold40 is more challenging and the accuracy of all methods tested is lower.  {\SubdivNet} again outperforms all mesh-based methods on Manifold40. However, the Transformer-based method PCT~\cite{journals/cvm/GuoPCT21} is more robust to distortion than the hierarchical networks. 

\begin{table}[t!]
    \centering
    \begin{tabu}{llrr}
    \toprule
        Method                                          & ModelNet40    & Manifold40\\
    \midrule
        PointNet++~\cite{DBLP:conf/cvpr/QiSMG17}        & 91.7\%        & 87.9\%\\
        PCT~\cite{journals/cvm/GuoPCT21}                & 93.2\%        & \textbf{92.4}\%\\
    \midrule
        SNGC~\cite{DBLP:conf/iccv/HaimSBML19}           & 91.6\%        & -\\
        MeshNet~\cite{DBLP:conf/aaai/FengFYZG19}        & 91.9\%        & 88.4\%\\
        MeshWalker~\cite{DBLP:journals/tog/LahavT20}    & 92.3\%        & 90.5\%\\
        SubdivNet (w/o majority voting)                 & -             & \textbf{91.2}\%\\
        SubdivNet                                       & -             & \textbf{91.5}\%\\
    \bottomrule
    \end{tabu}
	\vspace{0.5em}
    \caption{Classification accuracy on ModelNet40~\cite{DBLP:conf/cvpr/WuSKYZTX15} and Manifold40. The first two rows are state-of-the-art point cloud methods with positions and normals as input. Other methods use meshes as input.}
    \label{tab:modelnet40}
\end{table}

\subsection{Segmentation}

In the mesh segmentation task, {\SubdivNet} is trained to predict labels for every face.
As remeshed input is used, the remeshed faces should be appropriately labeled before we can start training.
To do so, we simply adopt a nearest-face strategy to build a mapping between the raw mesh and the remeshed one. 

\subsubsection{Human Body Segmentation} The human body dataset, labeled by~\cite{DBLP:journals/tog/MaronGATDYKL17}, contains 381 training shapes from SCAPE~\cite{DBLP:journals/tog/AnguelovSKTRD05}, FAUST~\cite{DBLP:conf/cvpr/Bogo0LB14}, MIT~\cite{DBLP:journals/tog/VlasicBMP08}, Adobe Fuse~\cite{adobefuse}, and 18 test shapes from SHREC07~\cite{giorgi2007shape}. 
The human body is divided into 8 segments. In this case, we used Liu et al.'s method~\cite{DBLP:journals/tog/LiuKCAJ20} to remesh the inputs to ensure lower distortion of details. 
Majority voting is employed in testing.
Table~\ref{tab:humanbody} gives the results, which shows that our  method  outperforms other methods. 

The standard input resolution of MeshCNN~\cite{DBLP:journals/tog/HanockaHFGFC19} is 1,500 faces. 
To find out whether the performance of MeshCNN is merely limited by the resolution, we additionally trained MeshCNN using 10,000-face inputs to enable a fair comparison. 

\begin{figure}[t!]
    \centering
    \includegraphics[width=\linewidth]{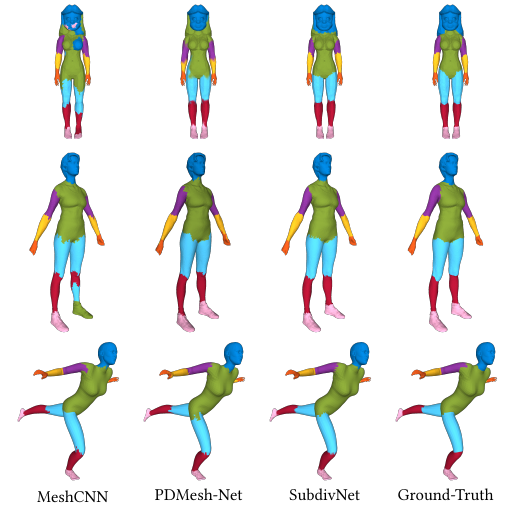}
    \caption{Segmentation results from the human body dataset~\cite{DBLP:journals/tog/MaronGATDYKL17}. {\SubdivNet} correctly classified all body parts, and gave more accurate boundaries.}
    \label{fig:human_seg}
\end{figure}

Some examples of the segmentation results are visualized in Fig.~\ref{fig:human_seg}. 
Compared to MeshCNN~\cite{DBLP:journals/tog/HanockaHFGFC19} and PD-MeshNet~\cite{DBLP:conf/nips/MilanoLR0C20}, {\SubdivNet}  more accurately segments parts, with more consistent boundaries.

\begin{table}[t!]
    \centering
    \begin{tabular}{@{}rlr@{}}
        \cmidrule[\heavyrulewidth]{2-3}
        & Method                                              & Accuracy \\ 
        \cmidrule{2-3}
        \ldelim\{{5}{.7cm}[Note A] \hspace{1.7em} & 
        Pointnet~\cite{DBLP:conf/cvpr/QiSMG17}              & 74.7\% \\
        & Pointnet++~\cite{DBLP:conf/nips/QiYSG17}           & 82.3\% \\
        & MeshCNN~\cite{DBLP:journals/tog/HanockaHFGFC19}    & 87.8\% \\
        & MeshCNN (10000 faces)                              & 65.3\% \\
        & PD-MeshNet~\cite{DBLP:conf/nips/MilanoLR0C20}      & 86.9\% \\
        & Toric Cover~\cite{DBLP:journals/tog/MaronGATDYKL17} & 88.0\% \\ 
        & SNGC~\cite{DBLP:conf/iccv/HaimSBML19}               & 91.3\% \\
        & PFCNN~\cite{DBLP:conf/cvpr/YangLPLT20}              & 91.5\% \\
        & MeshWalker~\cite{DBLP:journals/tog/LahavT20}        & 92.7\% \\
        & DiffusionNet~\cite{sharp2021diffusionnet}           & 91.7\% \\
        & SubdivNet (w/o majority voting)                     & 91.1\% \\
        & SubdivNet                                           & \textbf{93.0}\% \\
        \cmidrule{2-3}
        \ldelim\{{4}{.7cm}[Note B] \hspace{1.7em} & MeshCNN~\cite{DBLP:journals/tog/HanockaHFGFC19}.    & 92.3\% \\
        & MeshWalker~\cite{DBLP:journals/tog/LahavT20}        & 94.8\% \\
        & DiffusionNet~\cite{sharp2021diffusionnet}           & 95.5\% \\
        & SubdivNet                                           & \textbf{96.6}\% \\
        \cmidrule{2-3}
        \ldelim\{{4}{.7cm}[Note C] \hspace{1.7em} & PD-MeshNet~\cite{DBLP:conf/nips/MilanoLR0C20}       & 85.6\% \\
        & HodgeNet~\cite{DBLP:journals/tog/0001021}           & 85.0\% \\
        & DiffusionNet~\cite{sharp2021diffusionnet}           & 90.3\% \\
        & SubdivNet                                           & \textbf{91.7}\% \\
        \cmidrule[\heavyrulewidth]{2-3} 
    \end{tabular}
	\vspace{0.5em}
    \caption{Mesh segmentation accuracy on the human body dataset~\cite{DBLP:journals/tog/MaronGATDYKL17}. Note A: these methods use different types of input data; results were projected onto the raw meshes. Notes B, C: these methods were evaluated on the edges and faces of simplified meshes, respectively. See Appendix \ref{app:metric} for  details of the projection and the metrics.}
    \label{tab:humanbody}
\end{table}

\subsubsection{COSEG}
\begin{figure}
    \centering
    \includegraphics[width=\linewidth]{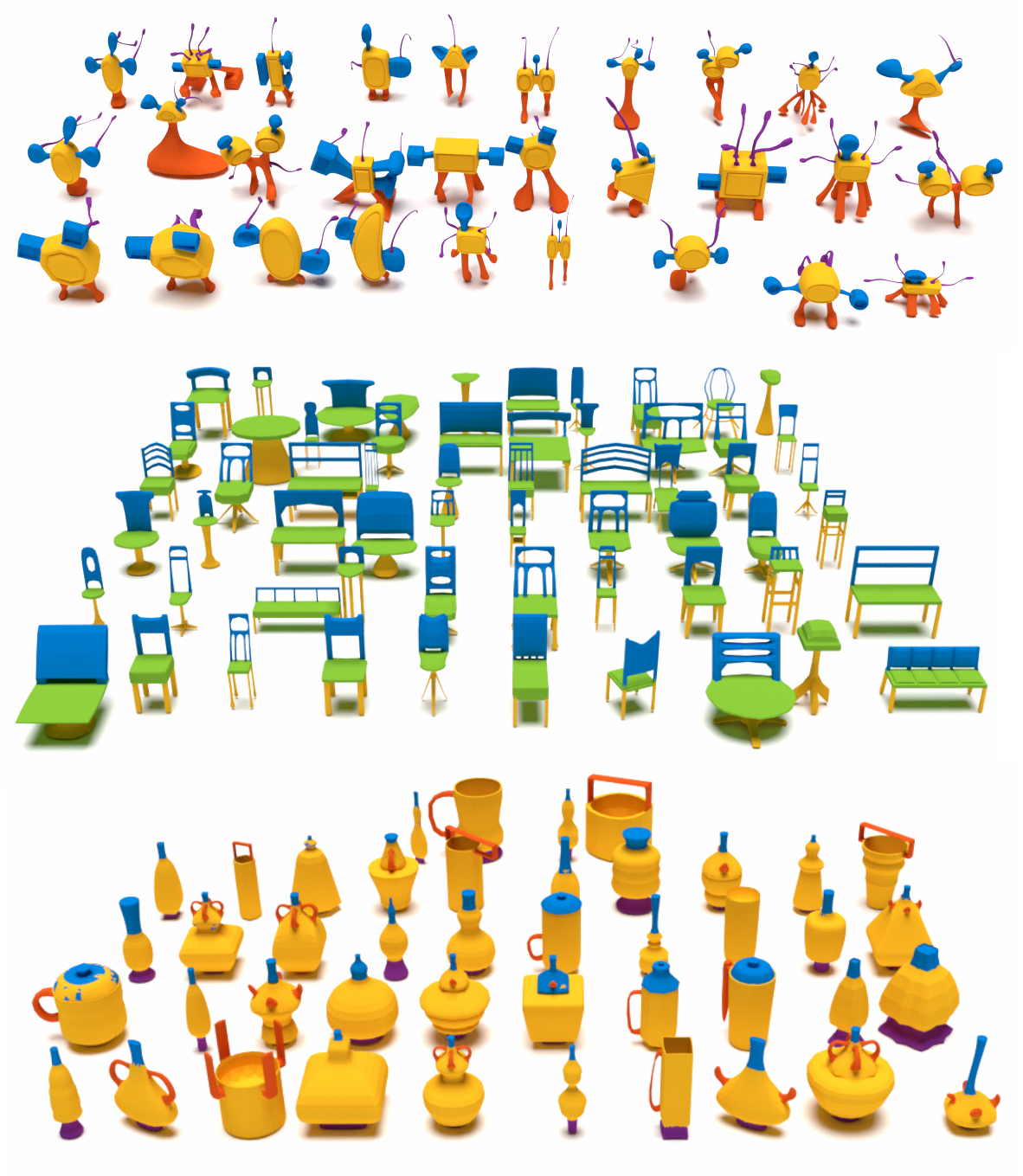}
    \caption{Gallery of segmentation results for the COSEG dataset.}
    \label{fig:coseg}
\end{figure}

We also assessed {\SubdivNet} on the three largest subsets of the COSEG shape dataset~\cite{DBLP:journals/tog/WangAK0CC12}: tele-aliens, chairs, and vases, which contain in turn 200, 400, and 300 models. They are segmented into only 3 or 4 parts, so we chose the MAPS algorithm as the remeshing method. We discovered that the MeshCNN-generated ids for chairs and vases do not match those in the original COSEG dataset. Therefore, we followed the training-testing split of MeshCNN for tele-aliens, but randomly split the training and tests set for chairs and vases in a 4:1 ratio.
{\SubdivNet} was trained on the three datasets independently. 
Quantitative results are provided in Table~\ref{tab:coseg} and example output is displayed in Fig.~\ref{fig:coseg}. 
Our method  achieves a significant improvement over MeshCNN~\cite{DBLP:journals/tog/HanockaHFGFC19} and PD-MeshNet~\cite{DBLP:conf/nips/MilanoLR0C20}. We also trained SubdivNet on another two random splits of vases and chairs. The mean accuracy and the standard deviation on the two datasets are $97.0\% \pm 0.6\%$ and $95.1\% \pm 1.5\% $, respectively. 

\begin{table}[t!]
    \centering
    \begin{tabular}{lrrrr}
        \toprule
        Method      & Vases & Chairs & Tele-aliens \\ 
       	\midrule
        MeshCNN~\cite{DBLP:journals/tog/HanockaHFGFC19} & 85.2\% & 92.8\% & 94.4\% \\
        PD-MeshNet~\cite{DBLP:conf/nips/MilanoLR0C20}   & 81.6\% & 90.0\% & 89.0\% \\
        SubdivNet   & \textbf{96.7}\% & \textbf{96.7}\% & \textbf{97.3}\% \\
        \bottomrule
    \end{tabular}
	\vspace{0.5em}
    \caption{Mesh segmentation accuracy on the COSEG dataset~\cite{DBLP:journals/tog/WangAK0CC12}. The training-testing split for vases and chairs differs from \cite{DBLP:journals/tog/HanockaHFGFC19}. Both MeshCNN and PD-MeshNet were trained on the new split, and  results were projected to the raw meshes in COSEG.}
    \label{tab:coseg}
\end{table}

\subsection{Shape Correspondence}
\begin{figure}
    \centering
    \includegraphics[width=\linewidth]{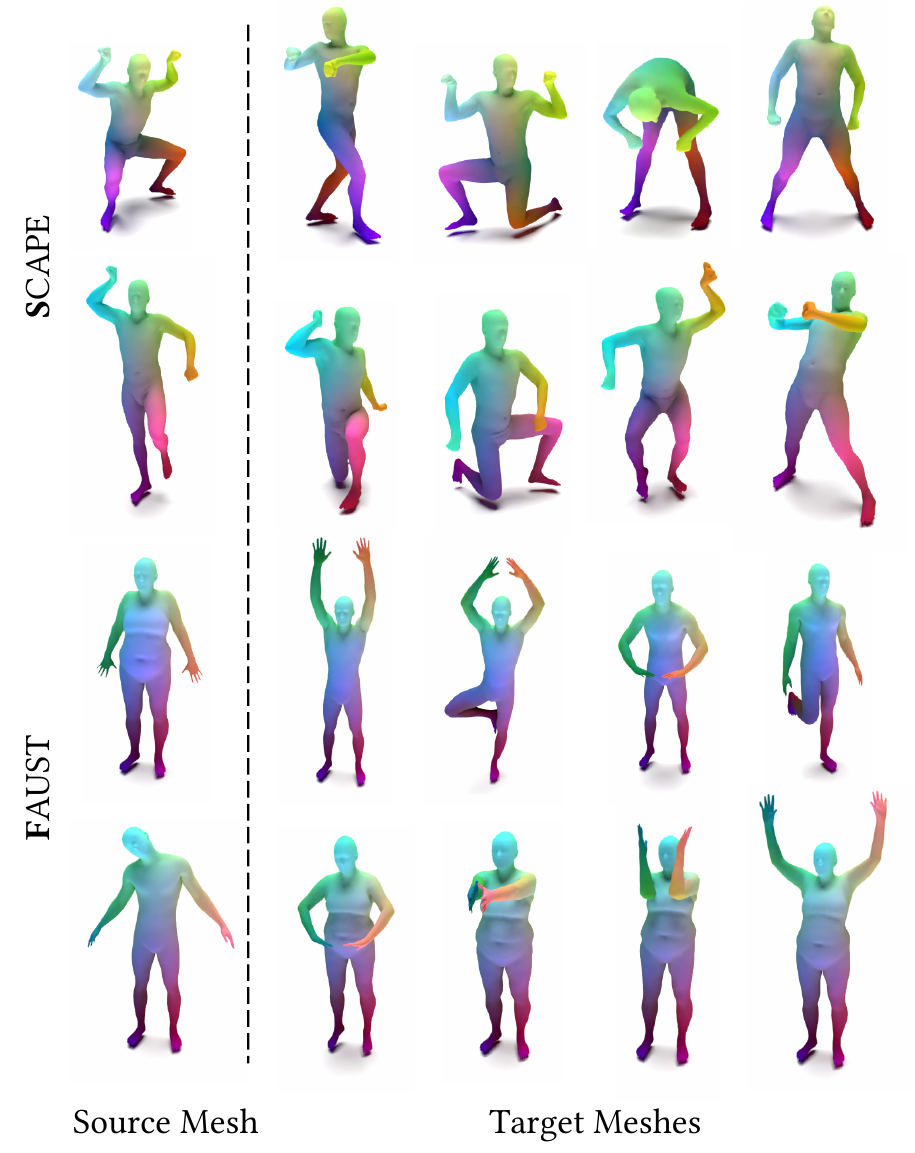}
    \caption{Shape correspondences learned using different datasets. Each row shows the output matching from the source mesh to the target meshes on the right. Corresponding positions share the same color.}
    \label{fig:correspondence}
\end{figure}

Our method can act as a robust feature extraction backbone for learning fine-grained shape correspondences between two meshes.
We demonstrate via human body matching using FAUST~\cite{DBLP:conf/cvpr/Bogo0LB14} and SCAPE~\cite{DBLP:journals/tog/AnguelovSKTRD05} datasets.
Specifically, our network was trained to predict 3-dimensional canonical human coordinates in a similar way to~\cite{mehta2017vnect}, but at mesh level.
The set of predicted coordinates is treated as an $\mathcal{R}^3$-valued function and the functional coordinates (dimension = 30) are computed based on the spectrum of the Laplace-Beltrami operator on the corresponding mesh.
We then build a functional map, a representation for non-rigid shape matching, between the source and target meshes, by solving a linear system as in~\cite{ovsjanikov2012functional}.
Lastly, the map is refined using ZoomOut~\cite{DBLP:journals/tog/MelziRRSWO19} and converted back to point-to-point correspondences.
Note that due to the scarcity of data, we additionally remeshed our geometries with different subdivision connectivities to augment   training data, which helps the model to learn tessellation-invariant robust features.

Our evaluation protocol follows Bogo et al.~\shortcite{DBLP:conf/cvpr/Bogo0LB14} and Donati et al.~\shortcite{donati2020deep}:
The shape correspondence error is calculated as the normalized mean geodesic distance between predicted and ground-truth mapped target positions on the target mesh. 
The datasets, FAUST and SCAPE, were respectively divided into 80:20 and 51:20 training/test splits.
Results from different combinations of the training and test sets are reported in Table~\ref{tab:correspondence} and results are visualized in Fig.~\ref{fig:correspondence}.
Our method achieves state-of-the-art matching and shows good generalizability across different datasets, demonstrating the effectiveness of the proposed method on this challenging task.

\begin{table}[t!]
    \centering
    \begin{tabular}{lcccc} 
    \toprule
    Method          & \textbf{F} & \textbf{S} & \textbf{F} on \textbf{S} & \textbf{S} on \textbf{F}  \\ 
    \midrule
    BCICP~\cite{DBLP:journals/tog/RenPWO18}   & 15.   & 16.   & -     & -     \\
    ZoomOut~\cite{DBLP:journals/tog/MelziRRSWO19} & 6.1   & 7.5   & -     & -     \\
    SURFMNet~\cite{DBLP:conf/iccv/RoufosseSO19} & 7.4  & 6.1   & 19.   & 23.   \\
    FMNet~\cite{DBLP:conf/iccv/LitanyRRBB17}   & 5.9   & 6.3   & 11.   & 14.   \\
    3D-CODED~\cite{DBLP:conf/eccv/GroueixFKRA18} & 2.5  & 31.   & 31.   & 33.   \\
    GeomFMaps~\cite{donati2020deep} & \textbf{1.9} & \textbf{3.0}   & \textbf{9.2}   & 4.3   \\
    \midrule
    Ours       & \textbf{1.9}   & \textbf{3.0}   & 10.5   & \textbf{2.6}  \\
    \bottomrule
    \end{tabular}
	\vspace{0.5em}
    \caption{Shape correspondence error ($\times$100) comparison.
    `\textbf{F}' and `\textbf{S}' indicate FAUST~\cite{DBLP:conf/cvpr/Bogo0LB14} and SCAPE~\cite{DBLP:journals/tog/AnguelovSKTRD05} datasets.
    `\textbf{F} on `\textbf{S}' means training on FAUST and testing on SCAPE and vice versa.
    }
    \label{tab:correspondence}
\end{table}

\subsection{Further Evaluation}
In this section, the key components of {\SubdivNet} are evaluated on SHREC11 (split 10) ~\cite{DBLP:journals/tog/WangAK0CC12} and the human body segmentation dataset~\cite{DBLP:journals/tog/MaronGATDYKL17}.

\subsubsection{Convolution Patterns and Network Architecture}
\label{sec:network}
\begin{table*}[t!]
    \centering
    \begin{tabular}{@{}rlcccr@{}}
        \cmidrule[\heavyrulewidth]{2-6}
        & Network Architecture & Kernel Size & Stride & Dilation   & Accuracy \\
        \cmidrule{2-6}
        \ldelim\{{2}{1.25cm}[Classification] \hspace{1.7em}
        & VGG-like               & 3 & 1 & 1 & 99.5\% \\
        & ResNet50 & 3,5 & 1,2 & 1 & 99.5\% \\
        \cmidrule{2-6}
        \ldelim\{{4}{1.3cm}[Segmentation] \hspace{1.7em}
        & U-Net~\cite{DBLP:conf/miccai/RonnebergerFB15} & 3     & 1 & 1  & 89.5\% \\
        & DeepLabv3+\cite{DBLP:conf/eccv/ChenZPSA18}   & 3, 5  & 1, 2 & 1,6,12,18  & 93.0\% \\
        & DeepLabv3+ w/o strided convolution        & 3     & 1 & 1,6,12,18  & 92.8\% \\
        & DeepLabv3+ w/o dilation         & 3, 5  & 1, 2 & 1  & 92.6\% \\
        \cmidrule[\heavyrulewidth]{2-6}
    \end{tabular}
	\vspace{0.5em}
    \caption{Ablation study on convolution kernel size, stride, dilation, and network architectures, evaluated on the SHREC11 (split 10, without voting)~\cite{DBLP:journals/tog/WangAK0CC12} and the human body segmentation dataset~\cite{DBLP:journals/tog/MaronGATDYKL17}. In the last two networks, the special convolutions are replaced by basic convolutions.}
    \label{tab:deeplab}
\end{table*}

Table 7 compares many networks with varied convolution patterns and architectures to see if 2D network architectures can aid 3D mesh learning. In 2D vision, the performance gap between segmentation networks is more significant than the classification networks (see the CityScape benchmark~\cite{Cordts2016Cityscapes} and ImageNet benchmark~\cite{DBLP:conf/cvpr/DengDSLL009}). Unsurprisingly, network architecture transfer is also more useful in 3D segmentation. Large kernel size and dilation are shown to be effective through the segmentation ablation studies. 

\subsubsection{Input Resolution}
\begin{table}[t!]
    \centering
    \begin{tabular}{cccr}
        \toprule
        Base Size   & Subdivision Depth & Faces & Accuracy \\ 
        \midrule
        48          & 3                 & 3072  & 99.3\% \\
        96          & 3                 & 6144  & 99.1\% \\
        48          & 4                 & 12288 & 99.5\% \\
        \bottomrule
    \end{tabular}
	\vspace{0.5em}
    \caption{Classification accuracy on SHREC11 (split 10)~\cite{DBLP:journals/tog/WangAK0CC12} using different input size.}
    \label{tab:resolution}
\end{table}

To examine the robustness of our network to input size, we tried several combinations of the base size and subdivision depth of inputs and retrained the network on \emph{Split10}. 
Table~\ref{tab:resolution} suggests that  performances are quite close. Because  remeshing distortion is more observable as the input size decreases, the results also show that our network can capture global shape and can tolerate input distortion to some extent. 

However, more input faces lead to a heavier network and more training time. On the other hand, if the input size is too small, e.g. a 48-face base mesh with one time subdivision, some categories are quite similar, such as the cats and the dogs. In this case, the network cannot distinguish them well.

\subsubsection{Input Features}
\begin{table}[t!]
    \centering
    \begin{tabular}{lr}
        \toprule
        Input                   & Accuracy  \\
        \midrule
        shape descriptor only  & 90.4\%    \\
        pose descriptor only   & 92.5\%    \\
        full input              & 93.0\%    \\
        \bottomrule
    \end{tabular}
    \vspace{0.5em}
    \caption{Ablation study on input features, evaluated on a human body segmentation dataset~\cite{DBLP:journals/tog/MaronGATDYKL17}. }
    \label{tab:input}
\end{table}

Unlike image pixels, which have identical size and shape everywhere,  shapes and sizes of mesh triangle faces represent local geometry. Thus, shape descriptors for the input features, i.e.\ areas, angles, and curvatures, are essential to the capability of {\SubdivNet}. Table~\ref{tab:input} indicates that both shape and pose are necessary for mesh learning.

\subsubsection{Learning on the Raw Meshes}
We implemented a variant of DeepLabv3+ with a feature propagation layer and additional convolution layers on the raw meshes. Table~\ref{tab:Propagation} shows that learning on the raw meshes slightly improves the segmentation quality. However, the extra layers require a 20\% increase in computing time. Details can be found in Appendix~\ref{app:network}.

\begin{table}[t!]
    \centering
    \begin{tabular}{lr}
        \toprule
        Method  & Accuracy \\ 
        \midrule
        DeepLabv3+ with feature propagation &  93.06\% \\ 
        DeepLabv3+ without feature propagation & 93.03\% \\
        \bottomrule
    \end{tabular}
	\vspace{0.5em}
    \caption{Ablation study for the feature propagation layer, based on the human body segmentation dataset~\cite{DBLP:journals/tog/MaronGATDYKL17}.}
    \label{tab:Propagation}
\end{table}

\subsubsection{Computation Time and Memory Consumption}
Our method was implemented with the Jittor deep learning framework~\cite{hu2020jittor}. Its flexibility enables efficient face neighbor indexing; convolution can be implemented with general matrix multiplication operators. As a result, the proposed network is as efficient as a 2D network.

We measured forward and backward propagation duration as well as GPU memory consumption on a 48-core CPU and a single TITAN RTX GPU. Table~\ref{tab:time_mem} shows that {\SubdivNet} is more than 20 times faster than an edge-based approach~\cite{DBLP:journals/tog/HanockaHFGFC19} using less than a third of the amount of GPU memory, achieving comparable performance to a highly optimized 2D CNN~\cite{DBLP:conf/eccv/ChenZPSA18}.

\begin{table}[t!]
    \centering
    \setlength{\tabcolsep}{3.0pt}
    \begin{tabular}{lrrr}
        \toprule
        Network & \begin{tabular}[c]{@{}c@{}}Input Size\\ (faces/pixels)\end{tabular} & \begin{tabular}[c]{@{}c@{}}Time\\ (ms)\end{tabular} & \begin{tabular}[c]{@{}c@{}}GPU Memory\\ (MB)\end{tabular} \\ 
        \midrule
        SubdivNet (DeepLabv3+) & 16384 & 47.4 & 1221 \\
        MeshCNN & 10000 & 1051.2 & 4090 \\
        2D DeepLabv3+   & 16384 & 20.1 & 612 \\
        \bottomrule
    \end{tabular}
	\vspace{0.5em}
    \caption{Computation time and GPU memory consumption of SubdivNet, MeshCNN~\cite{DBLP:journals/tog/HanockaHFGFC19}, and the 2D DeepLabv3+~\cite{DBLP:conf/eccv/ChenZPSA18}. The number of layers in DeepLabv3+ was reduced to be the same as for SubdivNet. Numbers are averaged over 1000 data samples.}
    \label{tab:time_mem}
\end{table}

\subsection{3D shape retrieval from the real-world}

The superior representation power of {\SubdivNet} allows us to effectively extract global shape descriptors for arbitrary 3D meshes.
We demonstrate this by retrieving 3D shapes from from partially-observed point clouds captured by an Asus Xtion Pro Live 3D sensor.
By jointly embedding both point clouds and meshes into the latent feature space, shape retrieval can be implemented as nearest neighbor search in the Euclidean-structured latent manifold.
Specifically, to build such a latent space, we first trained a denoising point cloud variational auto-encoder using the encoder architecture from \cite{DBLP:conf/cvpr/QiSMG17} and the decoder network from \cite{fan2017point}. The point clouds were synthesized from the mesh dataset with virtual cameras.
Then we extracted the bottleneck features (dimension = 32) corresponding to all the point clouds in our dataset and used them to directly supervise SubdivNet, obtaining a mapping from  mesh space to the latent space.

\begin{figure}
    \centering
    \includegraphics[width=\linewidth,trim={0 0 10em 0},clip]{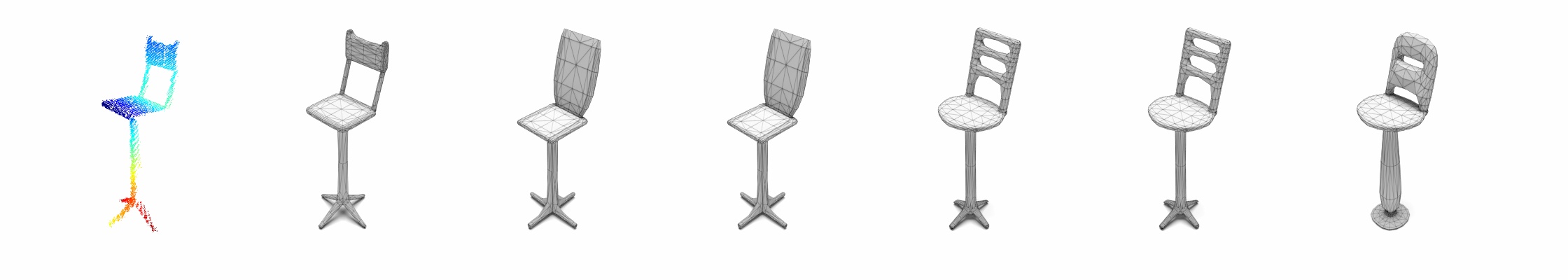}
    \includegraphics[width=\linewidth,trim={0 0 10em 0},clip]{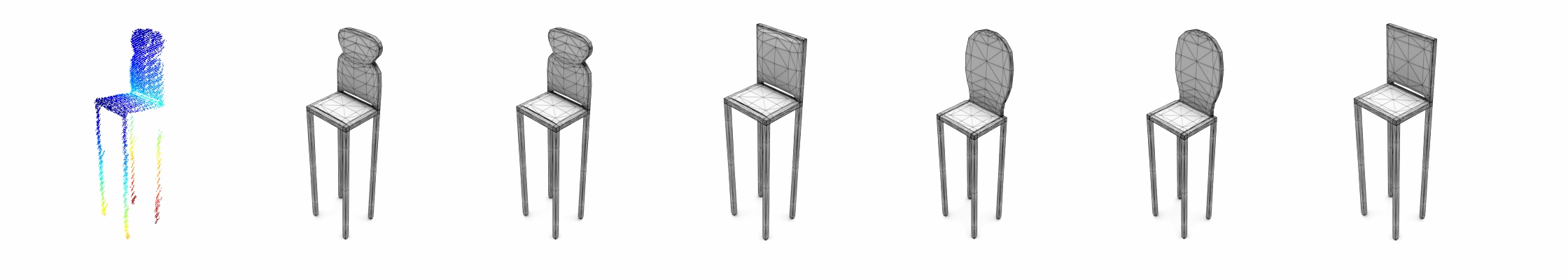}
    \includegraphics[width=\linewidth,trim={0 0 10em 0},clip]{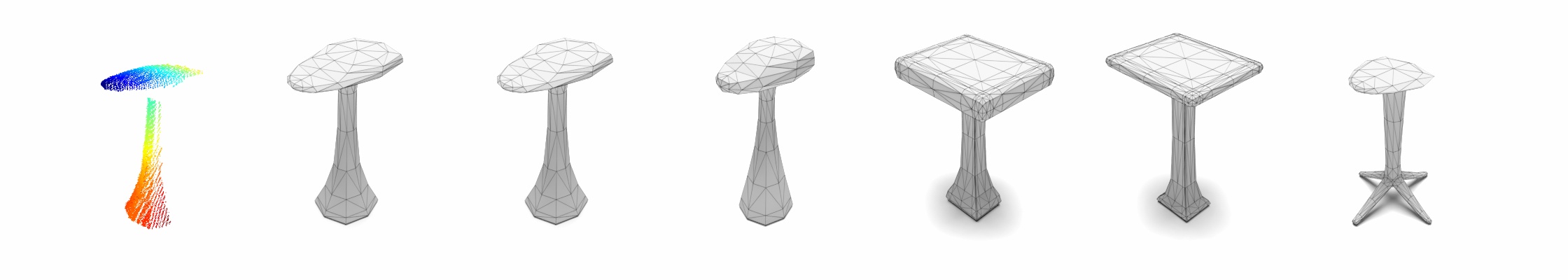}
    \includegraphics[width=\linewidth,trim={0 0 10em 0},clip]{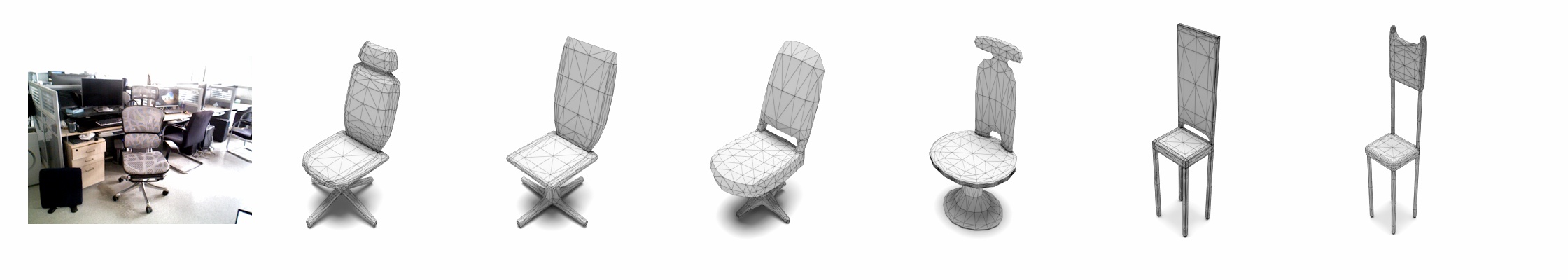}
    \includegraphics[width=\linewidth,trim={0 0 10em 0},clip]{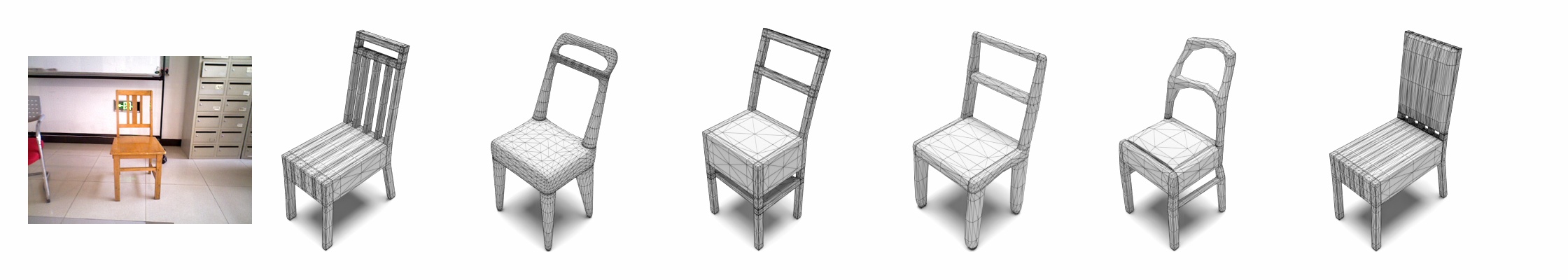}
    \includegraphics[width=\linewidth,trim={0 0 10em 0},clip]{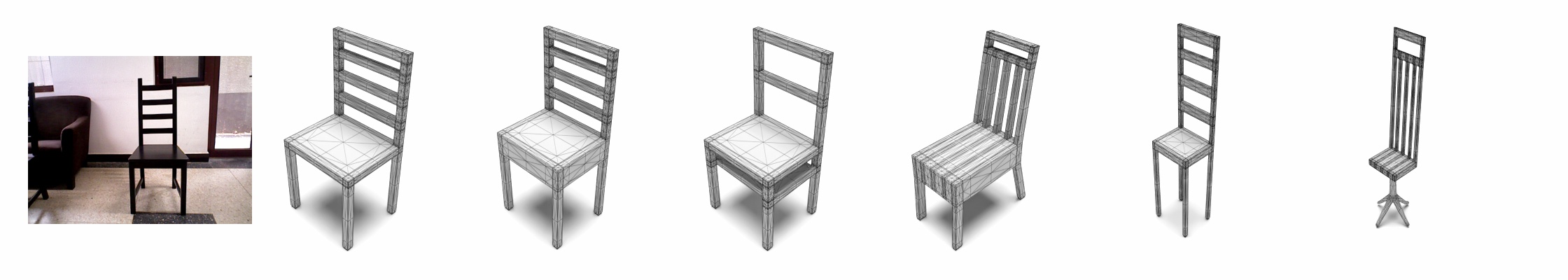}
    \caption{Shape retrieval results using our method. Left: input point clouds. Other columns: retrieved results. The last three rows are from real-world RGB-D captures, where only the back-projected depth points are used as input. Retrieval results are ordered from left to right by similarity in feature space.}
    \label{fig:retrieval}
\end{figure}

The chair models from the COSEG dataset~\cite{DBLP:journals/tog/WangAK0CC12} were used to train our network. Evaluation on the synthetic point clouds gives top1, top5, and top10 recall rates of 76.8\%, 83.3\%, and 88.0\%, respectively. Fig.~\ref{fig:retrieval} shows retrieval results for both synthetic point clouds and real-world depth scans.

\section{Limitations and Future Work}

\subsection{Convolution}
The origins of convolutions in 2D CNNs can be traced back to signal filtering. 
In our framework, we must take care to ensure that convolution is ordering-invariant, so we first transform the neighborhood features, rather than apply direct signal convolutions, leading to an isotropic kernel for all neighboring faces.
Another consequence is that the number of convolution parameters does not increase with kernel size as for 2D convolutions. One possible strategy is to differentiate faces by their distances from the center.
 
\subsection{Subdivision Connectivity}
Subdivision connectivity plays a crucial role in {\SubdivNet}, providing a uniform feature aggregation scheme from local to global. We believe it is a key factor in our method surpassing other mesh learning methods. However, remeshing is necessary to apply our method to an arbitrary mesh. 
As discussed in Sec.~\ref{sec:meshgen}, both remeshing approaches used have limitations, with a trade-off between mesh quality and base mesh size. 
Processing imperfect meshes, polygon soups, objects with borders, and large-scale scenes with flaws requires more thought. Adaptive remeshing~\cite{DBLP:conf/siggraph/LeeSSCD98} may be more helpful than the current uniform remeshing.
We also note that majority voting improves the final performance, showing that the differences between the remeshed shape and the raw mesh affect the results to some extent. 
Better methods of remeshing~\cite{DBLP:journals/tog/SharpSC19a} are needed, or better, ways of downsampling without needing remeshing at all.

\subsection{Applications}
This paper demonstrates SubdivNet's effectiveness on single shape analysis, but the current network cannot be directly applied on large-scale scenes due to the limitation of the remeshing technique. Yet the proposed convolution is promising for being integrated into a scene network, because it does not rely on subdivision connectivity. Apart from the applications in the paper, our ideas could also potentially be applied to traditional geometric problems, such as mesh smoothing and denoising, deformation, registration of multiple meshes, etc. They could also be employed in specific areas that require human knowledge or professional skill. 
For example, we could learn the natural right way up for a mesh, or  choose the best orientation of a mesh for 3D printing~\cite{DBLP:journals/cg/EzairME15}.

\section{Conclusions}

This work has presented a novel deep learning framework, {\SubdivNet}, for 3D geometric learning on meshes. 
The core of {\SubdivNet} is a general and flexible mesh convolution using a mesh pyramid structure for effective feature aggregation.
We first utilize self-parameterization to remesh the input mesh to have Loop subdivision sequence connectivity. 
That allows a well-defined, uniform mesh hierarchy to be constructed over the input shape. 
We then use mesh convolution operators which support user-specified kernel size, stride, and dilation. 
Pooling and upsampling are also naturally supported by subdivision connectivity. 
This enables the direct application of well-known 2D image CNNs to mesh learning. Our evaluations indicate that {\SubdivNet} surpasses existing mesh learning approaches in both accuracy and efficiency.

\begin{acks}
This work was supported by the Natural Science Foundation of China (Project Number 61521002) , Research Grant of Beijing Higher Institution Engineering Research Center and Tsinghua-Tencent Joint Laboratory for Internet Innovation Technology.
\end{acks}

\bibliographystyle{ACM-Reference-Format}
\bibliography{bibliography}

\appendix

\section{Network Implementation}
\label{app:network}
Here, we show in detail how the proposed mesh convolution defined above can be integrated into 2D network architectures to provide solutions for general 3D tasks such as mesh classification and segmentation. The code is publicly available at \url{https://xxx.xxx.xxx}.

\emph{Convolution Neighborhood Indexing.} When the kernel size $> 3$, the neighborhood can be found by depth-first search (DFS). Because a face is exactly adjacent to three neighbors, the search process results in a binary DFS tree. The rearrangement is then obtained by in-order traversal of the binary tree. As neighbors can be indexed in parallel, the process is efficient.

\emph{Classification Network.} We implemented a VGG-like network, which simply has two blocks of basic convolution, batch normalization, and ReLU layers at each resolution. Max-pooling is use for downsampling. Experimentally, we find this simple convolutional network provides sufficient performance. Thus, we do not choose a more sophisticated architecture, e.g.\ ResNet. 

\begin{figure}[t!]
    \centering
    \includegraphics[width=\linewidth]{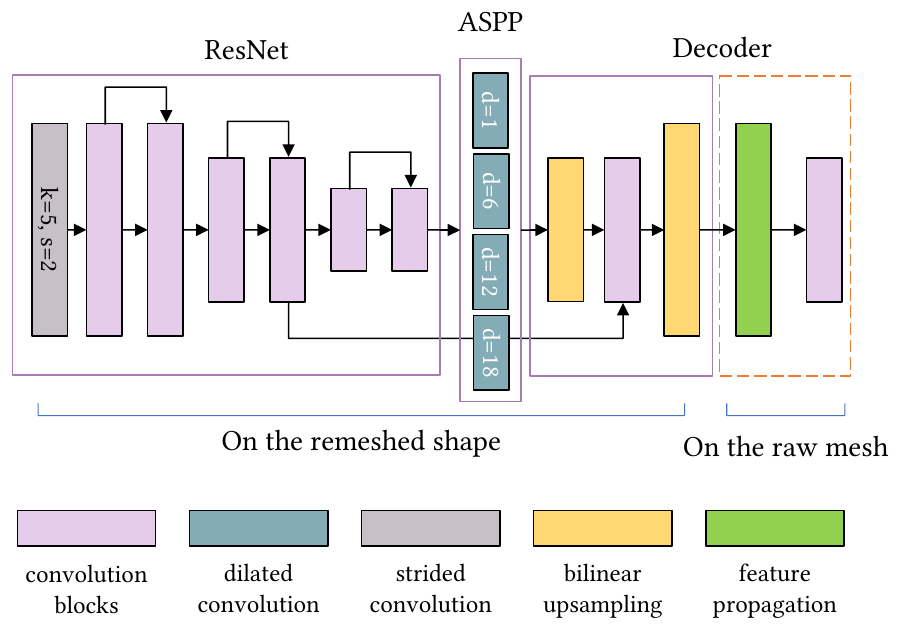}
    \caption{DeepLabv3+ architecture. If the layers in the dashed orange box are enabled, the network can output per-face features on the raw mesh.}
    \label{fig:deeplabv3+}
\end{figure}

\emph{DeepLabv3+}. Because the raw training meshes (before remeshing) typically have far fewer triangles than pixels in a 2D image, we simply use ResNet50~\cite{ResNet16} as a feature extractor instead of xception~\cite{DBLP:conf/cvpr/Chollet17}, both for efficiency and to avoid overfitting. The kernel size and stride of the first convolution are also lowered to 5 and 2, respectively, and we reduce the number of downsampling layers to 3. One key component of DeepLabv3+ is the atrous spatial pyramid pooling (ASPP) that stacks multiple dilated convolutions to enlarge the receptive field. We use dilations in our experiments of 1, 6, 12, and 18. The purple boxes in Fig.~\ref{fig:deeplabv3+} depict the network architecture.

\emph{DeepLabv3+ with Feature Propagation.} With the feature propagation layer and convolutions on the raw mesh (see the dashed orange box in Fig.~\ref{fig:deeplabv3+}), we can achieve a complete end-to-end pipeline. The extra layers increase computation by about 20\%. 

\emph{Input shape.} For the classification task, the base mesh size is 48, similar to the $7 \times 7$ output feature map of the last convolution in VGG and ResNet. The subdivision depth is set to 4, resulting in the finest mesh having 12288 faces. For dense prediction tasks, the base size and subdivision depth are 256 and 3, respectively, reaching a similar number of faces of raw meshes in human body segmentation to balance the prediction quality and computational efficiency. We find these hyper-parameters work well for evaluation on public datasets. However, for meshes with higher resolution, one may choose a larger subdivision depth for better results.

\emph{Training.} The Adam optimizer is employed to train the networks. In all  experiments, we trained the network 4 times on the training set and reported the best results on the test set.

\section{Further Remeshing Details}
\label{app:remeshing}
To construct the subdivision sequence connectivity, the MAPS algorithm~\cite{DBLP:conf/siggraph/LeeSSCD98} establishes a bijective map between the raw mesh and the decimated version. 
In detail, MAPS iteratively removes the maximum independent set of vertices. 
When a vertex is removed, MAPS first re-triangulates the 1-ring neighbors, and calculates a conformal map over the local region between the before and after states. 
The removed vertices are also parameterized on the decimated mesh. 
After the raw mesh has been simplified onto a base mesh, a global parameterization is constructed. 
Then Loop subdivision~\cite{loop1987smooth} without vertex update is applied to the base mesh $d$ times, where $d$ is the subdivision depth, and the vertices of the subdivided mesh are projected onto the raw mesh using the global parameterization. Because the global parameterization links the raw mesh and the simplified mesh, this idea is also called \emph{self-parameterization}. One obvious advantage of MAPS is that it supports any genus as long as the decimation process does not break the topology. 

Recently, Liu et al.~\shortcite{DBLP:journals/tog/LiuKCAJ20} proposed a modified MAPS method, utilizing edge collapse based decimation, e.g.\ qslim~\cite{DBLP:conf/siggraph/GarlandH97}, rather than vertex removal, which improves the decimation quality.

In practice, we find both methods have limitations. In MAPS, the order of vertex removal is crucial. For example, repeated removal of limb vertices in  a horse mesh will lead to insufficient sampling of the limbs in the output, and ultimately the hooves cannot be fully reconstructed. 
This causes significant distortion if the mesh contains small but important details (see Fig.~\ref{fig:remeshing}). 
Liu et al.~\shortcite{DBLP:journals/tog/LiuKCAJ20} tackle the issue with a better decimation algorithm and prohibit collapses that cause poor triangle quality. 
However, doing so restricts the lowest base size the algorithm can reach (see Fig.~\ref{fig:failure_liu}). 

\begin{figure}[t!]
    \centering
    \includegraphics[width=\linewidth]{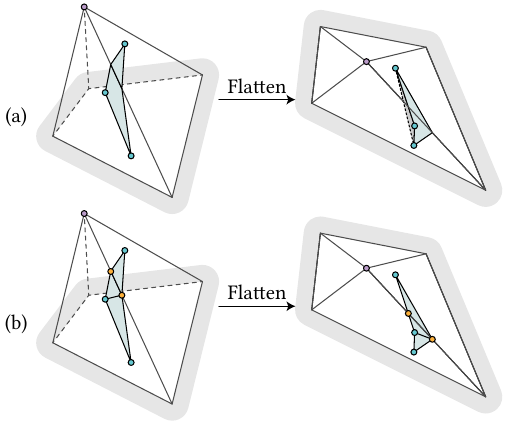}
    \caption{$UV$ Flip. (a) Example of $UV$ flip after local parameterization. The three blue vertices form a face in the input mesh,  already removed in earlier steps. Now they are parameterized in the current simplified mesh. (b). Splitting the blue triangle into three triangles with orange vertices prevents the problem.}
    \label{fig:uv_flip}
\end{figure}

More importantly,  \emph{$UV$ flip} may occur, so that a triangle face on the original mesh cannot be mapped to a triangular region in the parameter domain. Sampling on flipped $UV$ triangles may lead to remeshing failure. In detail, when removing a vertex, MAPS first flattens the vertex and its one-ring neighbors. The flattening, or local parameterization, may be invalid because of $UV$ flip (Fig.~\ref{fig:uv_flip}(a) shows a case). Because global parameterization is the composition of a sequence of local parameterizations,  $UV$ triangle flip is more probable when the base size is lower. Fortunately, $UV$ flip does not occur if the three vertices lie on the same face of the current decimated mesh. 
Thus, to avoid $UV$ flip, during decimation and parameterization, we split problematic triangles along with the triangulation of the current decimated mesh. 
For example, in Fig.~\ref{fig:uv_flip}(b), the triangle with three blue vertices is divided into three smaller triangles as it crosses an edge of the simplified mesh.

However, Liu et al.~\shortcite{DBLP:journals/tog/LiuKCAJ20} do not solve this issue. Instead, if $UV$ flip occurs after collapsing an edge and parameterizing the related region, they simply abandon this edge collapse operation and move to the next candidate edge to be collapsed. 
However, as more vertices of the raw mesh are parameterized to the simplified mesh, $UV$ flip may become inevitable if the input mesh is over-decimated. 
However, in 2D CNNs, the size of the feature map of the last layer is often very small, e.g.\ $7\times7$ pixels in ResNet~\cite{ResNet16}. Thus, for some inputs, Liu et al.'s method cannot meet our requirements.

\begin{table*}[!htb]
	\centering
	\begin{tabular}{lrrrr}
			\toprule
			Method      & HumanBody  & COSEG Vases & COSEG Chairs & COSEG Tele-aliens\\
			\midrule
			MeshCNN~\cite{DBLP:journals/tog/HanockaHFGFC19} & 92.3\% & 92.7\% & 98.1\%  & 97.6\% \\
			MeshWalker~\cite{DBLP:journals/tog/LahavT20}    & 94.8\% & - & - & 99.1\% \\
			SubdivNet                                       & \textbf{96.6}\% & \textbf{98.1}\% & \textbf{99.5}\% & \textbf{99.4}\% \\
			\bottomrule
	\end{tabular}
	\vspace{0.5em}
	\caption{Mesh segmentation accuracy using MeshCNN's metric. Vases and chairs use a train-test split different from that originally used by MeshCNN. }
	\label{tab:segmeshcnn}
\end{table*}

\section{Manifold40}
\label{app:manifold40}

\begin{figure}
    \centering
    \includegraphics[width=\linewidth]{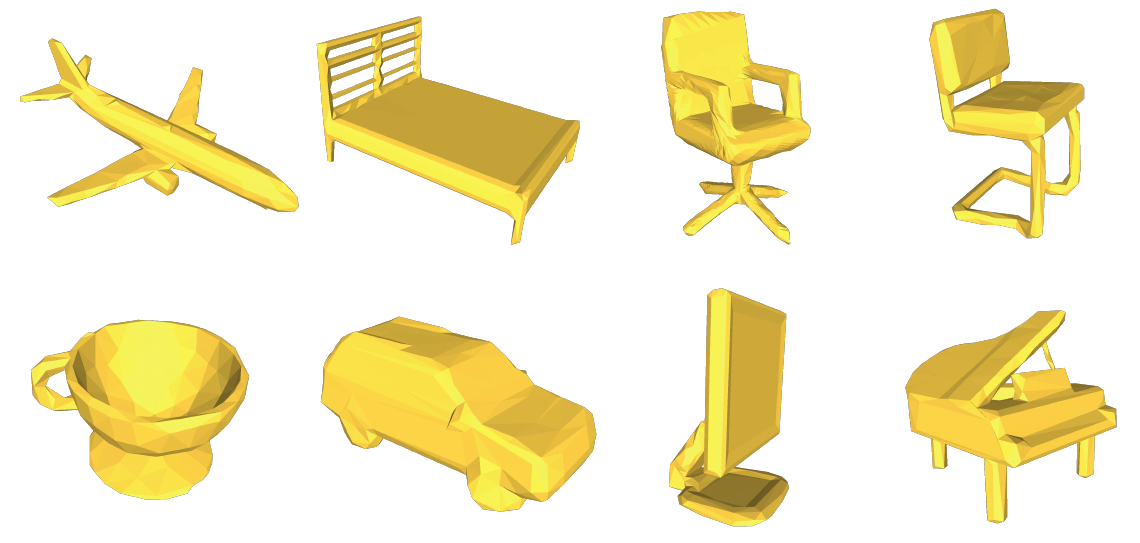}
    \caption{Examples from Manifold40.}
    \label{fig:manifold40}
\end{figure}

We employed a reconstruction algorithm~\cite{DBLP:journals/corr/abs-1802-01698} to make the models in ModelNet40 watertight. This method constructs an octree for the shape, extracts isosurfaces, and projects vertices onto the original shape. Small and isolated components are cleaned after reconstruction. Sometimes, we found that  non-manifold vertices may occur; we split these vertices using MeshLab~\cite{DBLP:conf/egItaly/CignoniCCDGR08}. Following the datasets contributed by~\cite{DBLP:journals/tog/HanockaHFGFC19}, all meshes were simplified to contain exactly 500 faces. Simplifying meshes also speeds up the remeshing process. Fig.~\ref{fig:manifold40} illustrates some examples from Manifold40.

We found some shapes in Manifold40  have a genus larger than 20. Because the decimation process of remeshing does not change the topology of shapes, it is almost impossible to reach a base size of 48 for such shapes. Therefore, we chose a more tolerant strategy: the base size of most meshes is enlarged to 96, and shapes with complicated topology are decimated as little as possible. As a result, except for 11 training samples, the base size of all other meshes was between 96 and 192. We simply discarded those 11 samples when training SubdivNet. To avoid heavy demands on computational resources, the depth was reduced to 3 from 4.

Using variable base sizes leads to variable input sizes. To incorporate them in a conventional batch-based training scheme, we padded meshes with empty faces to ensure all inputs have the same  size in a mini-batch. Because the global pooling layer after convolutions does not restrict the mesh to have a fixed number of faces, {\SubdivNet} can be trained and evaluated with variable input sizes.

\section{Evaluation Metrics for Mesh Segmentation}
\label{app:metric}

In the segmentation experiments, different inputs and evaluation metrics are employed by the approaches compared. The original human body segmentation contains up to 30k faces in a mesh. However, in COSEG, the number of faces ranges from hundreds to thousands. Both human body segmentation and COSEG datasets offer per-face labels on meshes. MeshCNN~\cite{DBLP:journals/tog/HanockaHFGFC19} simplifies the meshes to  1500 faces, and projects the segmentations onto the edges on simplified meshes. PD-MeshNet~\cite{DBLP:conf/nips/MilanoLR0C20} and HodgeNet~\cite{DBLP:journals/tog/0001021} also employ  simplified datasets but use face labels. To fairly compare {\SubdivNet} with other approaches, we report the performance of {\SubdivNet} using three  evaluation metrics.

\emph{Per-face Accuracy}. The metric is the overall accuracy on faces of the original meshes before simplification. It is also the default metric presented in this paper. For rows marked by Note A in Table~\ref{tab:humanbody}, we mapped other forms of segmentations to the original meshes. In detail, for point cloud methods~\cite{DBLP:conf/cvpr/QiSMG17,DBLP:conf/nips/QiYSG17}, we uniformly sampled 4096 points on the mesh surface. Then the segmentations on point clouds are projected to faces on meshes by finding the nearest point. For PD-MeshNet, the face label on the original mesh is obtained by finding the nearest face center in the simplified mesh. For MeshCNN, we first generate per face segmentation on the simplified meshes with edge labels, weighed by edge lengths. Then the nearest query strategy is used.

\emph{Per-edge Soft Accuracy on Simplified Meshes}. The metric is from MeshCNN, which collects the overall accuracy on edges with a soft criterion: an edge's prediction is true if it equals any of its neighbor's ground truth labels.  We directly projected our segmentation results to the edges of MeshCNN’s simplified meshes by querying the nearest edge center. Accuracy is calculated by public code for MeshCNN. Rows marked Note B in Table~\ref{tab:humanbody} use this metric. Table~\ref{tab:segmeshcnn} presents an additional experiment on COSEG under this metric.

\emph{Per-face Hard Accuracy on Simplified Meshes}. The metric is the overall accuracy on faces of the simplified meshes. The segmentation results of {\SubdivNet} are also directly projected onto the simplified meshes. Rows marked Note C in Table~\ref{tab:humanbody} use this metric.

\end{document}